\documentclass{article}
\usepackage{graphicx} 
\usepackage{amsmath} 
\usepackage{amsfonts} 
\usepackage{mathrsfs}
\usepackage{caption}
\usepackage{verbatim}
\usepackage{subcaption}
\usepackage{diagbox}
\usepackage{authblk}
\usepackage{hyperref}

\begin{document}
\title{Recovering the state and dynamics of autonomous system with partial states solution using neural networks}
\author{\large{Vijay Kag}\thanks{fixed-term.Vijay.Kag@in.bosch.com}}
\affil{\small{Bosch Research and Technology Center, Bangalore}}
\date{\small{March 2024}}

\maketitle
\section{Introduction}
Autonomous system is a system which does not explicitly depend on its independent variable i.e.  time \cite{strogatz2000nonlinear}. The dynamics of states are dependent on states itself.
Such systems can be found in nature and have applications in modeling chemical concentrations, population dynamics, n-body problems in physics etc.  One can express such system in form of these set of governing equations,

 \begin{equation}
     \begin{split}
         &\frac{d x_1}{dt} = f_1(x_1,x_2,...,x_k) \\
         &\frac{d x_2}{dt} = f_2(x_1,x_2,...,x_k)  \\
         &\frac{d x_k}{dt} = f_k(x_1,x_2,...,x_k)  
     \end{split}
 \end{equation}
 
In this work we are going to see how we can obtain dynamics of states based on solution of limited partial states. The proposed method can find the state and dynamics of which the data is provided in the training, although we do not claim to accurately find the solution of states whose data is not utilized while training. For this we leverage the method known as deep hidden physics model as described in \cite{Raissi_dhp_2018}.

\newpage 

\section{Deep hidden physics model}
Deep hidden physics models as described in \cite{Raissi_dhp_2018} can be used to discover underlying physical laws based on scattered data coming from experiments or simulations. However it learns the physics in a black box way, we do not have explicit representation of terms in equations. The method is as follows, it consists of two neural network, one network predicts the solution $u$ and other network predicts the hidden dynamics $\mathcal{N}$. Let's consider the governing equation as given form in \ref{nonlin_eqn}

\begin{equation}
\frac{\partial{u}}{\partial{t}} = \mathcal{N}(t,x,u,u_x,u_{xx},..) 
\label{nonlin_eqn}
\end{equation}

for $x \in \Omega, t \in [0,T], u: \Omega \times t  \rightarrow \mathbb{R} $ \\
%\newpage
we define residual $r(x,t)$ as, 

\begin{equation}
r(x,t) = \frac{\partial{u}}{\partial{t}} - \mathcal{N}
\label{residual_eqn}
\end{equation}

The arguments passed in $\mathcal{N}$ are known as candidate functions, depend on the nature of problem and domain knowledge, one can pass the list of possible candidate functions. 
We define loss function for data and equation as,
\begin{equation}
    \mathcal{L}_{data} = \frac{1}{N} \sum_{i=1}^{N} |u(x_i,t_i)-u^*(x_i,t_i)|^2
\end{equation}

\begin{equation}
    \mathcal{L}_{eqn} = \frac{1}{M} \sum_{j=1}^{M} |r(x_j,t_j)|^2
\end{equation}

where $\{x_i, t_i, u^*(x_i,t_i)\}$ are set of data points for reference. This data could be coming from experiments or simulations and  $\{x_j, t_j\}$ are set of points known as collocation points to constrain the physical law. Depending on the problem and amount of data, these points can be either be considered same as given data points or can be sampled randomly from the domain. 
We construct total loss functions as,

\begin{equation}
    \mathcal{L}_{total} = \mathcal{L}_{data} + \mathcal{L}_{eqn}
\end{equation}
The parameters of these two neural networks can be learned by minimizing the total loss function 
$\mathcal{L}_{total}$. Once the model is trained properly, one network predicts the solution and other predicts the underlying hidden terms of governing equation in complete black box form, we do not know the functional form of it.  
\newpage
\section{Illustrative problems}{\label{problem_setup}}
Let’s consider 2D autonomous system given by,

\begin{equation}
     \begin{split}
         &\frac{d x_1}{dt} =  \dot{x}_1 = f_1(x_1,x_2)    \\
         &\frac{d x_2}{dt} =  \dot{x}_2 = f_2(x_1,x_2)
     \end{split}
 \end{equation}
 
Given some sparse set of data in time for states $x_1, x_2$. We are interested in finding the states and dynamics $\dot{x}_1$, $\dot{x}_2$ using deep hidden physics model. We proceed with constructing neural network 
$\mathscr{N}_s$ which takes input as $[t]$ and outputs states $[\mathscr{N}_{x_1} ,\mathscr{N}_{x_2}]$ and second neural network $\mathscr{N}_f$ which takes input as $[\mathscr{N}_{x_1} ,\mathscr{N}_{x_2}]$ and outputs dynamics $[ \mathscr{N}_{\dot{x}_1},\mathscr{N}_{\dot{x}_2}]$.
We construct the losses based on data and equations, First we define individual data losses as,

\begin{equation}
\mathcal{L}_{d,1} = \lVert x_1 - \mathscr{N}_{x_1}\rVert_2,
\end{equation}

\begin{equation}
\mathcal{L}_{d,2} = \lVert x_2 - \mathscr{N}_{x_2}\rVert_2,
\end{equation}

Next we define loss equation as,

\begin{equation}
\mathcal{L}_{eq,1} = \lVert \frac{d \mathscr{N}_{x_1}}{dt}  - \mathscr{N}_{\dot{x}_1} \rVert_2,
\end{equation}

\begin{equation}
\mathcal{L}_{eq,2} = \lVert \frac{d \mathscr{N}_{x_2}}{dt}  - \mathscr{N}_{\dot{x}_2} \rVert_2,
\end{equation}

where $\lVert q \rVert_2 = \frac{1}{N} \sum_{i=1}^{N} |q_i|^2 $  defines mean squared error.\\

We define net data loss as,
\begin{equation}
\mathcal{L}_{d} = \alpha_1 \mathcal{L}_{d,1} + \alpha_2  \mathcal{L}_{d,2}
\end{equation}

We define net equation loss as,
\begin{equation}
\mathcal{L}_{eq} = \mathcal{L}_{eq,1} + \mathcal{L}_{eq,2}
\end{equation}

The total loss given by,

\begin{equation}
\mathcal{L}_{total} = \mathcal{L}_{d} + \mathcal{L}_{eq}
\end{equation}

The parameters of both the neural networks $\mathscr{N}_s$ ,  $\mathscr{N}_f$ are learned simultaneously by minimizing total loss $\mathcal{L}_{total}$. For optimization, We use Adam optimizer which performs gradient based optimization \cite{adam_opt}. The other training parameters are the number of data points, collocation points, epochs, learning rate etc. We trained the model until the $\mathcal{L}_{total}$ converges to $\mathcal{O}(10^{-4})$

We study the model based on different type of functions $f_1,f_2$ and contribution of individual data losses which is done by choosing $\alpha_1,\alpha_2 \in \{0,1\}$. 
To generate reference data we solve ODEs using scipy odeint package. 

\subsection{Type-a system}
\begin{figure}
     \centering
     \begin{subfigure}[b]{0.45\textwidth}
         \centering
         \includegraphics[width=\textwidth]{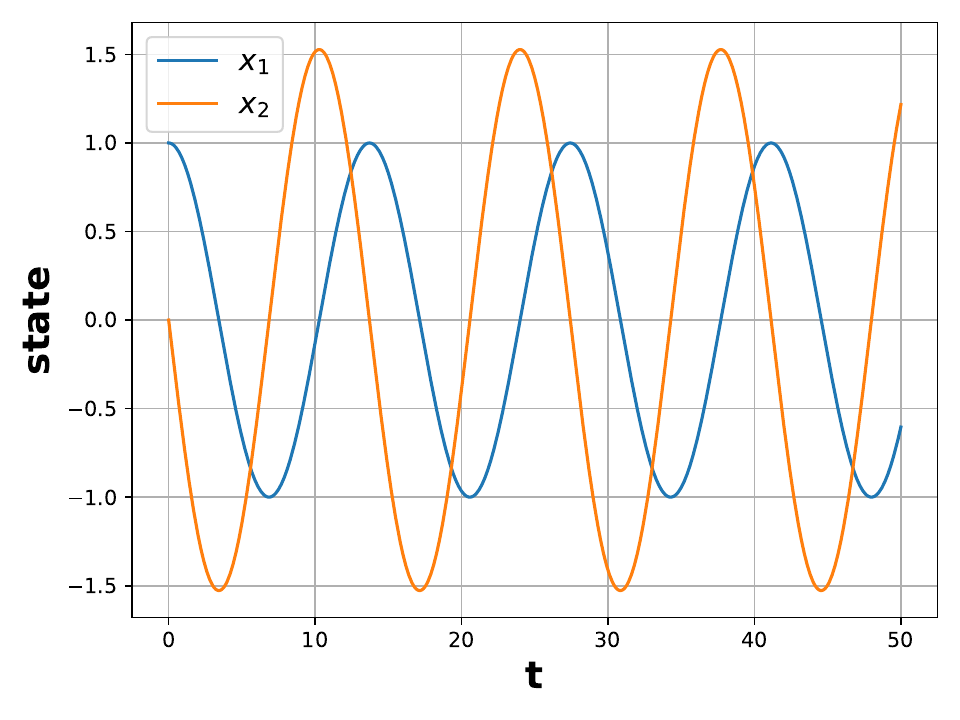}
         \caption{evolution of states}
         %\label{}
     \end{subfigure}
     \hfill
     \begin{subfigure}[b]{0.45\textwidth}
         \centering
         \includegraphics[width=\textwidth]{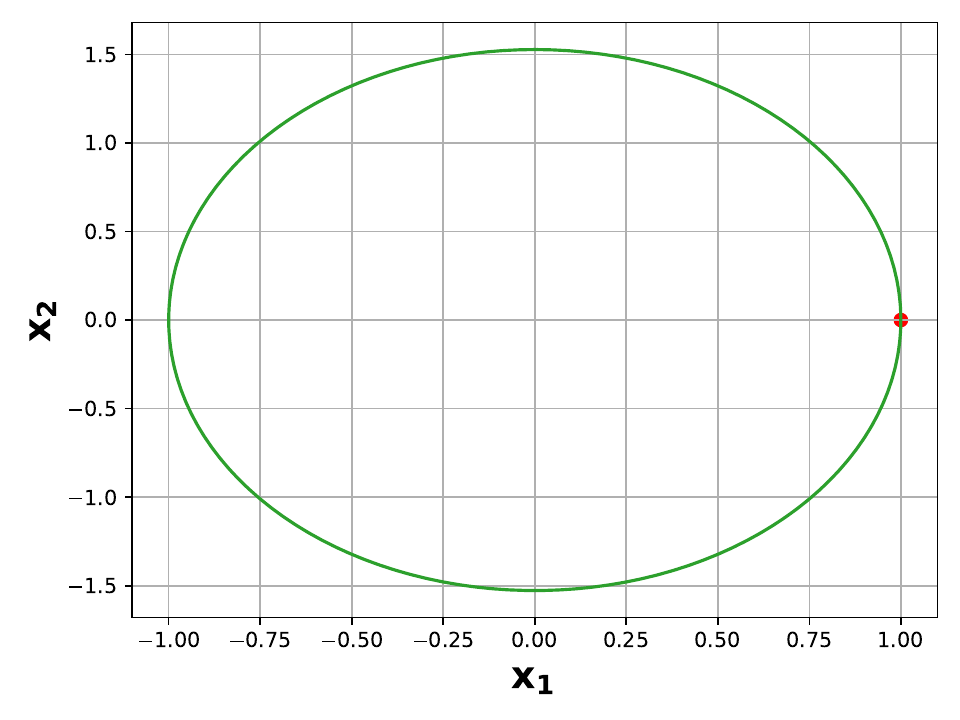}
         \caption{phase space}
         %\label{}
     \end{subfigure}
        \caption{Type-a system, $f_1 = 0.3 x_2, \  f_2 = -0.7 x_1$}
        \label{type1_sol}
\end{figure}

{\label{Type_a_func_section}}
We consider Type-a system by,

\begin{equation}
     \begin{split}
         & f_1 = 0.3 x_2   \\
         & f_2 = -0.7 x_1
     \end{split}
 \end{equation}
 
with initial condition as $x_1(0) = 1 , x_2(0) =0 $. The system is solved upto t = 50 unit. The nature of solution is shown in figure \ref{type1_sol} 

We train deep hidden physics model using 501 data points from reference solution of $x_1, x_2$
and additionally we consider 1000 random collocation points which are sampled using latin hypercube sampling (LHS) \cite{LHS} at each gradient descent step. The network architecture for $\mathscr{N}_s$ consists of 2 hidden layers with 64 neurons each and network $\mathscr{N}_f$ consists of 2 hidden layers with 64 neurons each. We train the model for 25000 epochs with learning rate of $10^{-3}$. The model is trained by considering states contributing to net data loss by choosing this set of values $\{ \alpha_1 = 1,\alpha_2 = 1 \}$, 
$\{ \alpha_1 = 1,\alpha_2 = 0 \}$ and $\{ \alpha_1 = 0,\alpha_2 = 1 \}$.
The results for this settings are described in the following subsections. To measure the deviation of predicted solution from reference we use metric known as Rel. $L_2$ Error which is defined as,

\begin{equation}\label{l2_error_def}
\text{Rel.} \ L_2 \ \text{Error} = \frac{\sqrt{\lVert x^* - x\rVert_2}} { \sqrt{\lVert x^* \rVert_2} }
\end{equation}
where $x^*, x $ is reference and predicted solution respectively.  
\newpage

\subsubsection{\label{type1_a1_1_a2_1} Case1: $\alpha_1 = 1,\alpha_2 = 1$ }
\begin{figure}
     \centering
     \begin{subfigure}[b]{0.45\textwidth}
         \centering
         \includegraphics[width=.95\linewidth]{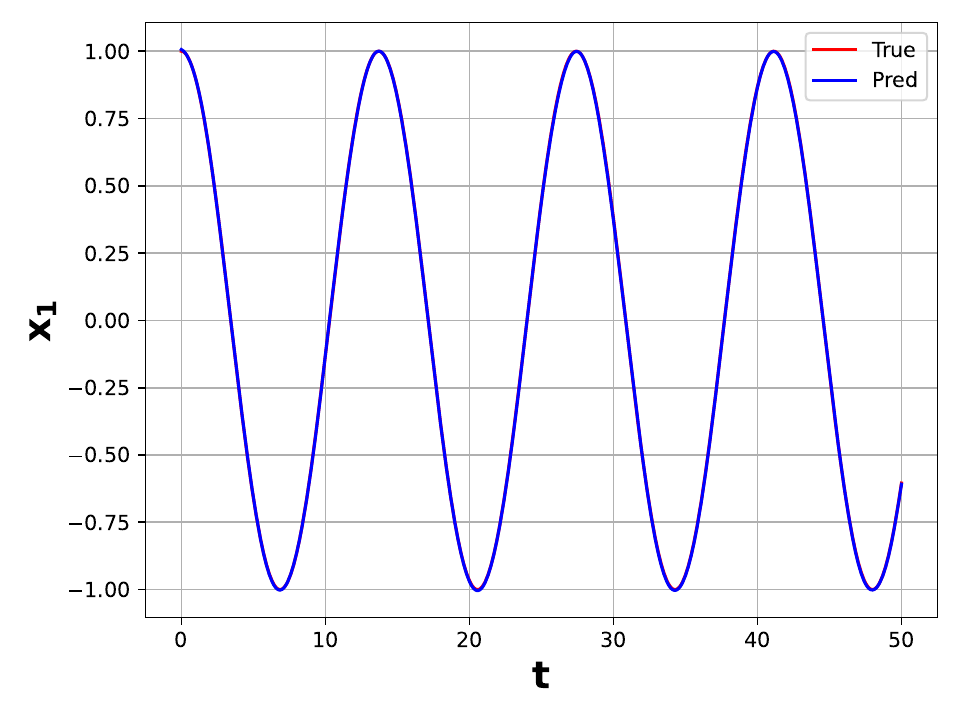}
         \caption{reference Vs predicted of state $x_1$ }
         %\label{}
     \end{subfigure}
     %\hfill
     \begin{subfigure}[b]{0.45\textwidth}
         \centering
         \includegraphics[width=.95\linewidth]{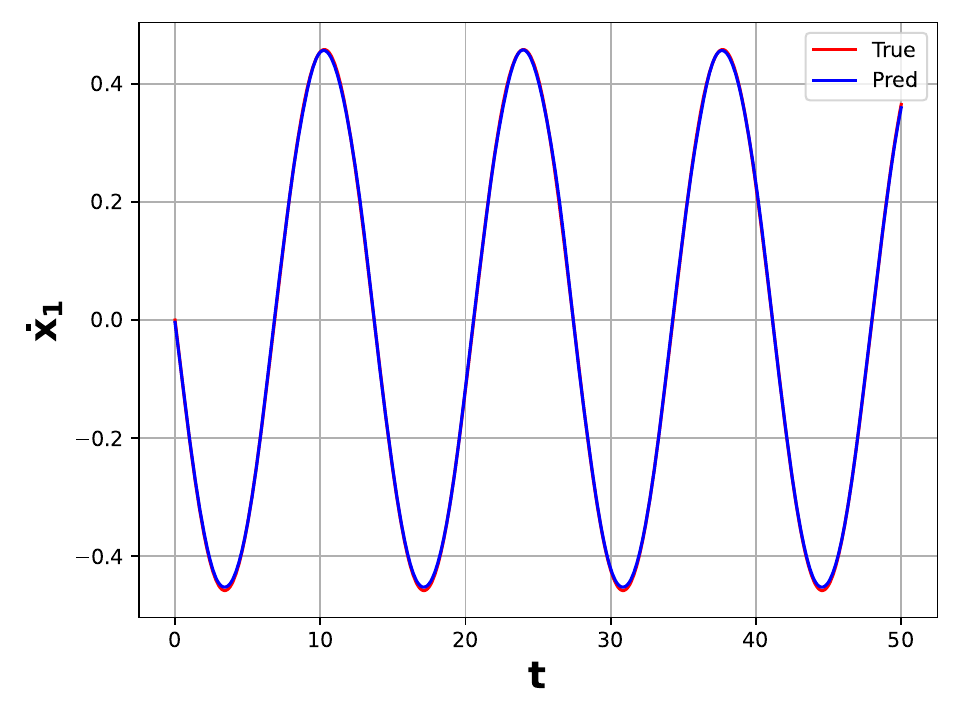}
         \caption{reference Vs predicted dynamics $\dot{x}_1$}
         %\label{}
     \end{subfigure}
     %\hfill
     \begin{subfigure}[b]{0.45\textwidth}
         \centering
         \includegraphics[width=.95\linewidth]{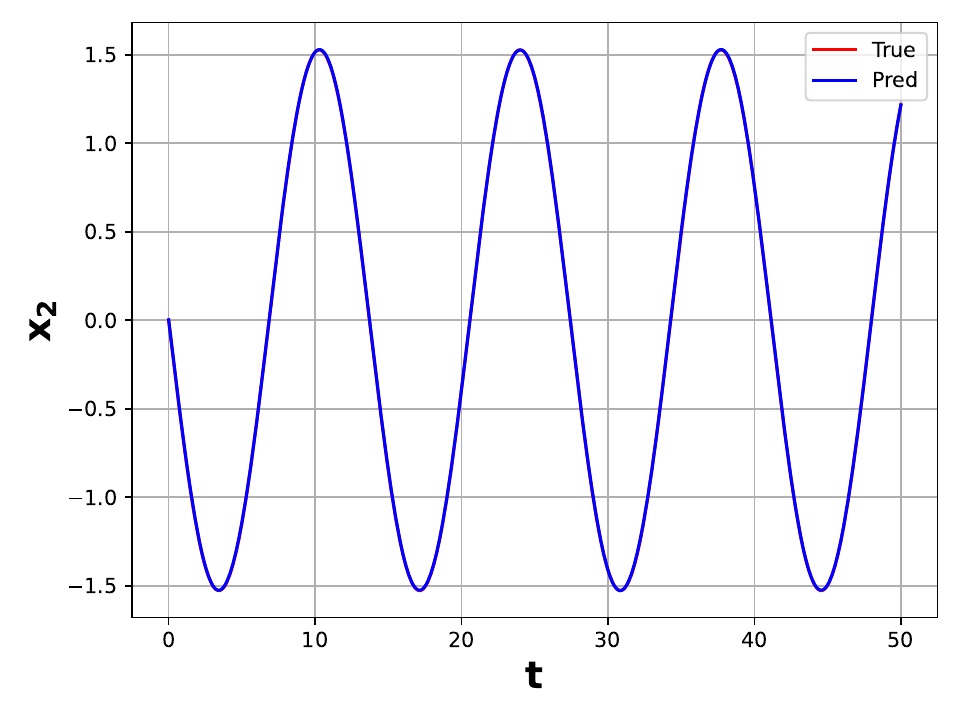}
         \caption{reference Vs predicted of state $x_2$}
         %\label{}
     \end{subfigure}
     %\hfill
     \begin{subfigure}[b]{0.45\textwidth}
         \centering
         \includegraphics[width=.95\linewidth]{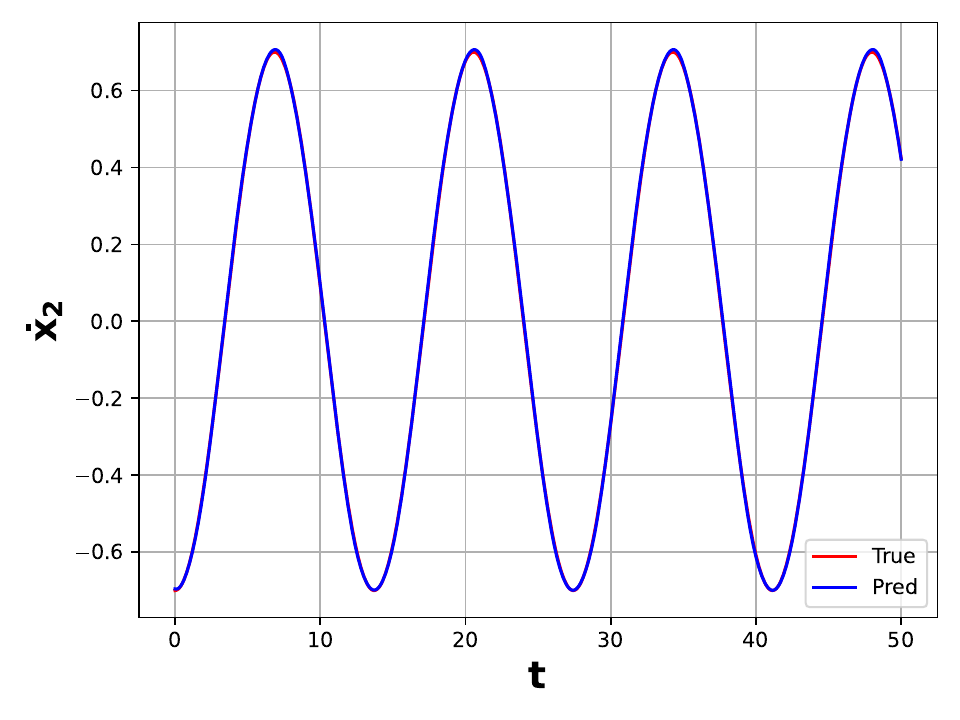}
         \caption{reference Vs predicted dynamics $\dot{x}_2$}
         %\label{}
     \end{subfigure}
        \caption{Type-a system, training with $\alpha_1 = 1,\alpha_2 = 1$  }
        \label{type1_a1_1_a2_1_sol}
\end{figure}

This shows that the data loss of both the states are contributing to net data loss. 
The results are shown in figure  \ref{type1_a1_1_a2_1_sol}

\begin{comment}
\begin{figure}
     \centering
     \begin{subfigure}[b]{0.45\textwidth}
         \centering
         \includegraphics[width=\textwidth]{x1_type1_a1_1_a2_1.pdf}
         \caption{reference Vs predicted of state $x_1$ }
         %\label{}
     \end{subfigure}
     \hfill
     \begin{subfigure}[b]{0.45\textwidth}
         \centering
         \includegraphics[width=\textwidth]{x2_type1_a1_1_a2_1.pdf}
         \caption{reference Vs predicted of state $x_2$}
         %\label{}
     \end{subfigure}
        \caption{Training with $\alpha_1 = 1,\alpha_2 = 1$  }
        \label{type1_a1_1_a2_1_sol}
\end{figure}
\end{comment}

\subsubsection{\label{type1_a1_1_a2_0} Case2: $\alpha_1 = 1,\alpha_2 = 0$ }
\begin{figure}
     \centering
     \begin{subfigure}[b]{0.45\textwidth}
         \centering
         \includegraphics[width=.95\linewidth]{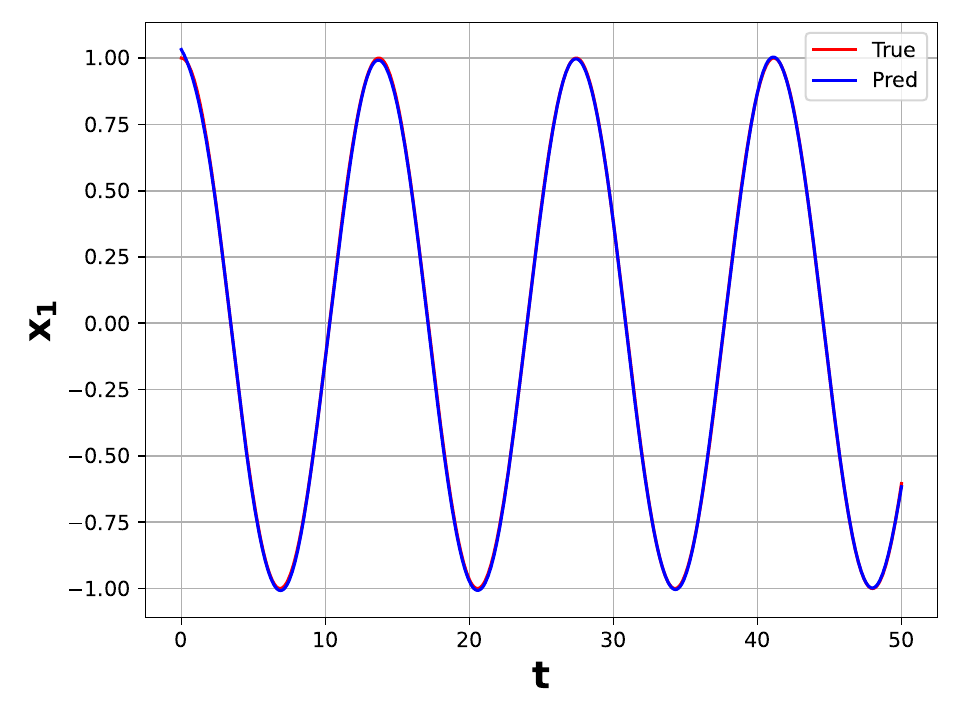}
         \caption{reference Vs predicted of state $x_1$ }
         %\label{}
     \end{subfigure}
     %\hfill
     \begin{subfigure}[b]{0.45\textwidth}
         \centering
         \includegraphics[width=.95\linewidth]{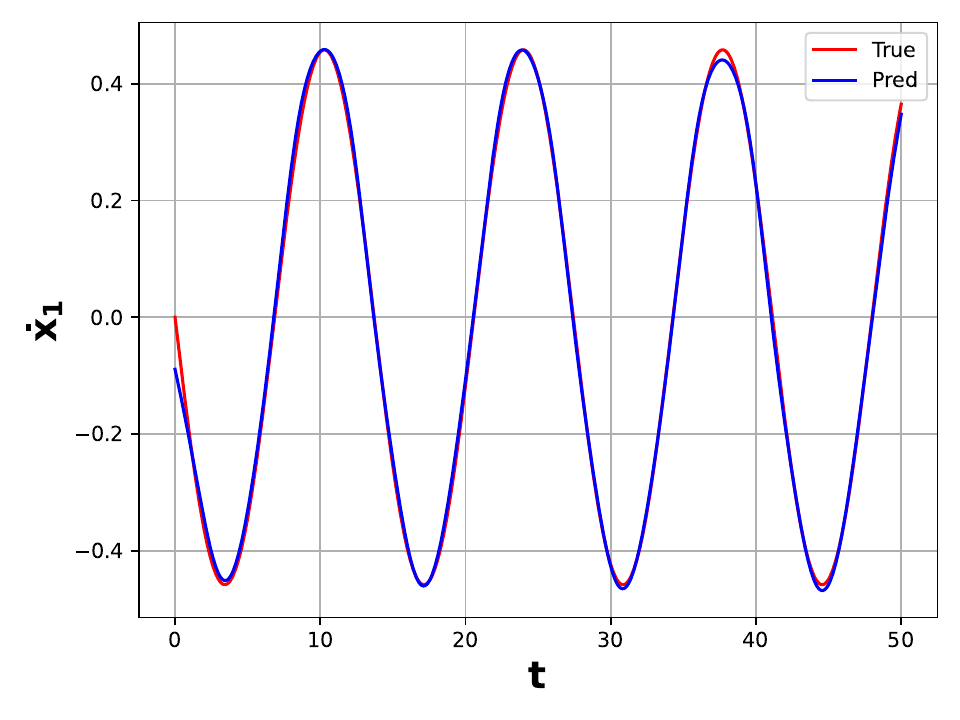}
         \caption{reference Vs predicted dynamics $\dot{x}_1$}
         %\label{}
     \end{subfigure}
      %\hfill
     \begin{subfigure}[b]{0.45\textwidth}
         \centering
         \includegraphics[width=.95\linewidth]{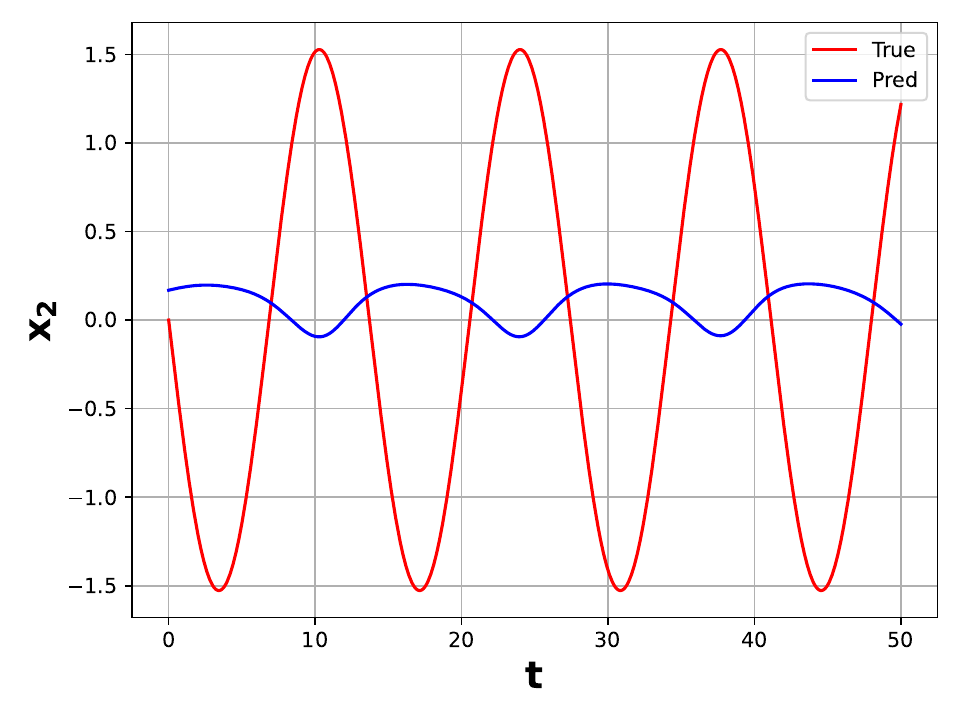}
         \caption{reference Vs predicted of state $x_2$}
         %\label{}
     \end{subfigure}
     %\hfill
     \begin{subfigure}[b]{0.45\textwidth}
         \centering
         \includegraphics[width=.95\linewidth]{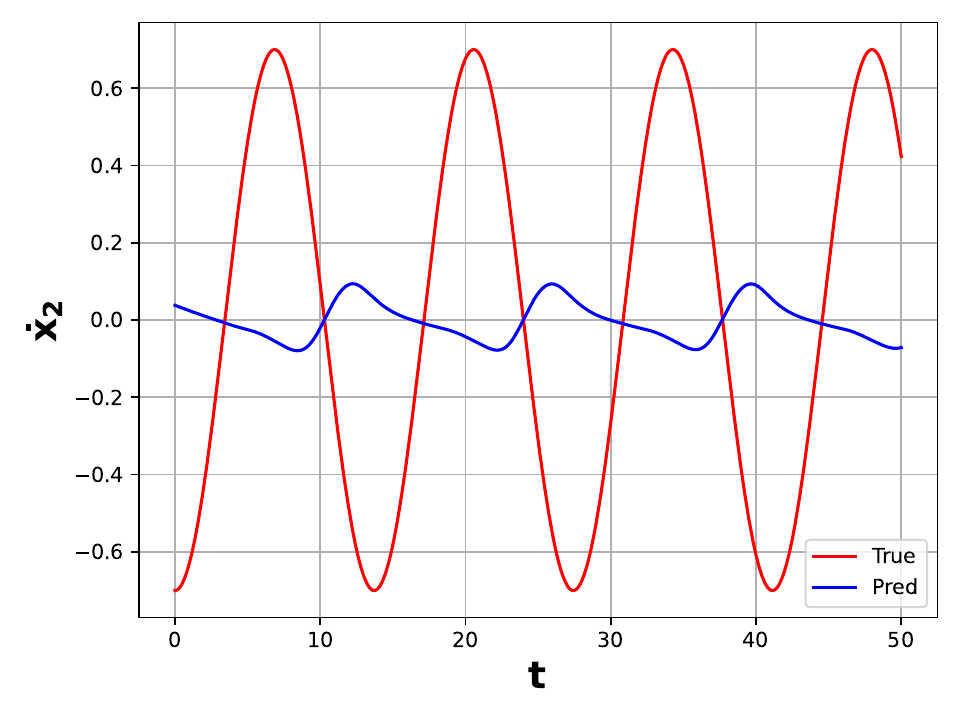}
         \caption{reference Vs predicted dynamics $\dot{x}_2$}
         %\label{}
     \end{subfigure}
        \caption{Type-a system, training with $\alpha_1 = 1,\alpha_2 = 0$  }
        \label{type1_a1_1_a2_0_sol}
\end{figure}

This shows that the data loss of only state $x_1$ is contributing to the net data loss. 
The results are shown in figure  \ref{type1_a1_1_a2_0_sol}

\subsubsection{\label{type1_a1_0_a2_1} Case3: $\alpha_1 = 0,\alpha_2 = 1$ }
\begin{figure}
     \centering
     \begin{subfigure}[b]{0.45\textwidth}
         \centering
         \includegraphics[width=.95\linewidth]{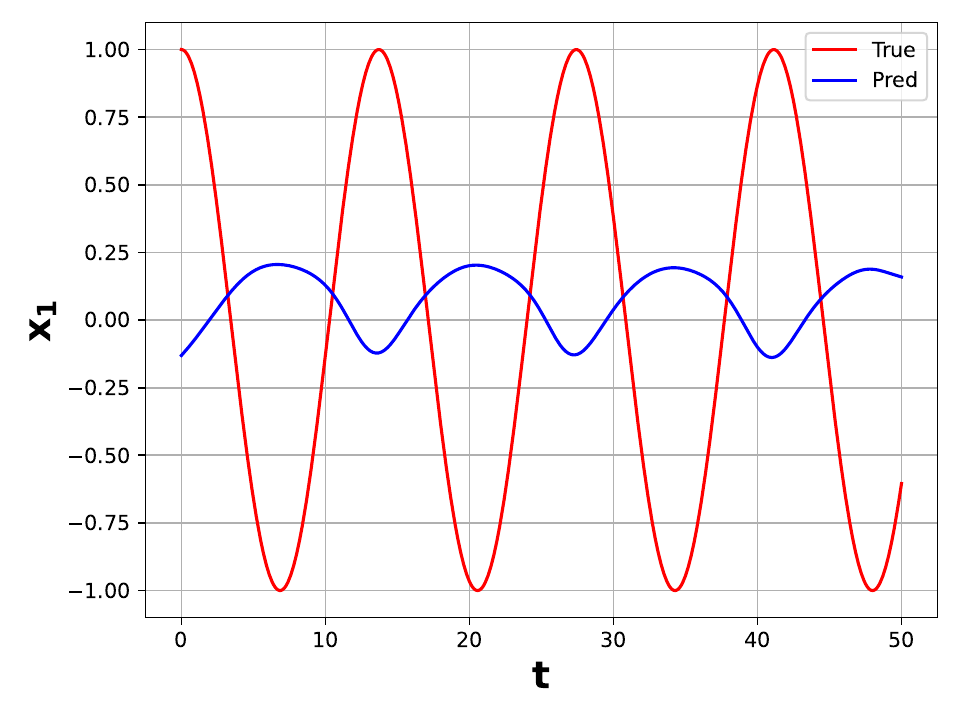}
         \caption{reference Vs predicted of state $x_1$ }
         %\label{}
     \end{subfigure}
     %\hfill
     \begin{subfigure}[b]{0.45\textwidth}
         \centering
         \includegraphics[width=.95\linewidth]{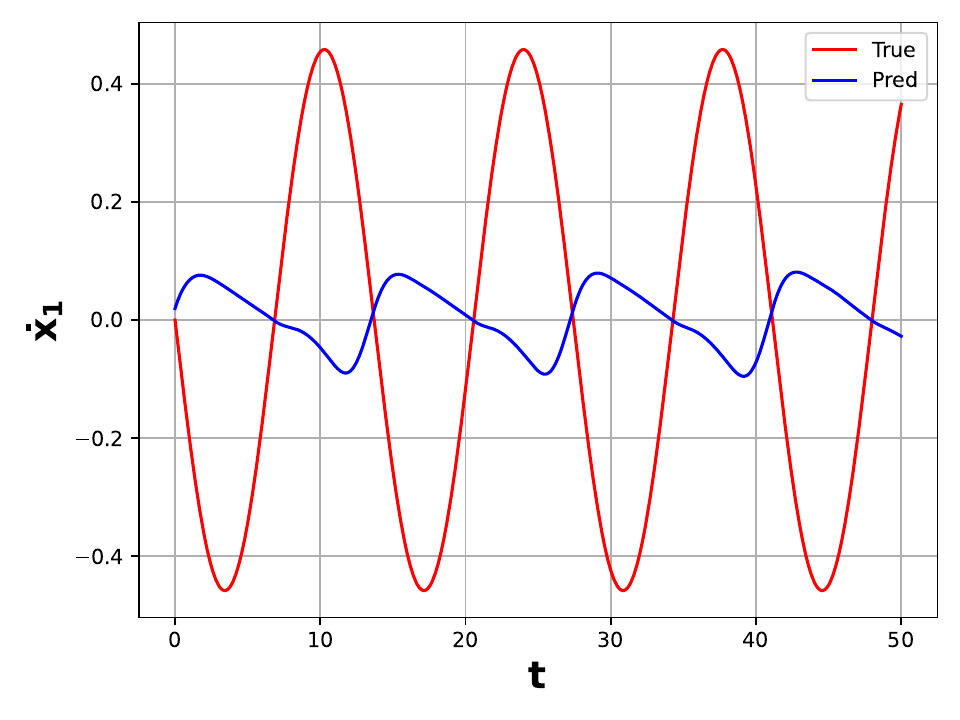}
         \caption{reference Vs predicted dynamics $\dot{x}_1$}
         %\label{}
     \end{subfigure}
     %\hfill
     \begin{subfigure}[b]{0.45\textwidth}
         \centering
         \includegraphics[width=.95\linewidth]{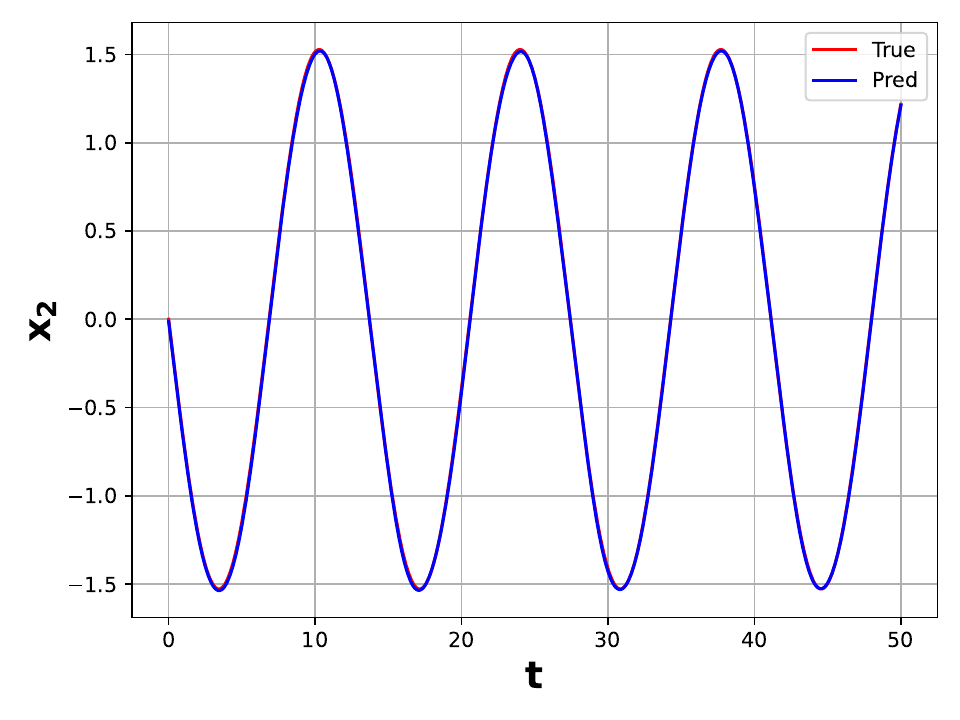}
         \caption{reference Vs predicted of state $x_2$}
         %\label{}
     \end{subfigure}
     %\hfill
     \begin{subfigure}[b]{0.45\textwidth}
         \centering
         \includegraphics[width=.95\linewidth]{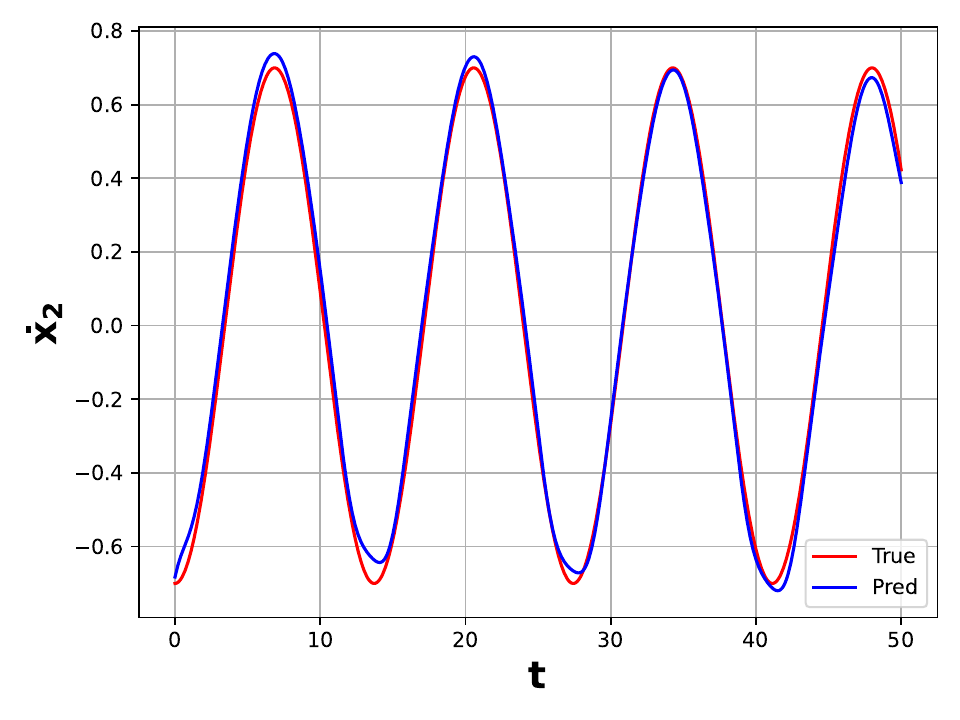}
         \caption{reference Vs predicted dynamics $\dot{x}_2$}
         %\label{}
     \end{subfigure}
        \caption{Type-a system, training with $\alpha_1 = 0,\alpha_2 = 1$}
        \label{type1_a1_0_a2_1_sol}
\end{figure}

This shows that the data loss of only state $x_2$ is contributing to the net data loss. The results are shown in figure  \ref{type1_a1_0_a2_1_sol}

{\renewcommand{\arraystretch}{1.2}
\begin{table}
\caption{$L_2$ Error for Type-a system with different contributions in data loss}
\label{type_a_error}
\begin{tabular}{|l||*{4}{c|}}\hline
%\backslashbox{$\alpha$}{quantities}
&\makebox[5.5em]{$x_1$}&\makebox[5.5em]{$\dot{x}_1$}&\makebox[5.5em]{$x_2$}
&\makebox[5.5em]{$\dot{x}_2$}\\\hline
$\alpha_1 = 1,\alpha_2 = 1$ & $3.90 \times 10^{-3}$ & $9.58 \times 10^{-3}$  & $2.60 \times 10^{-3}$  & $8.03\times 10^{-3}$  \\\hline
$\alpha_1 = 1,\alpha_2 = 0$ & $9.10 \times 10^{-3}$ & $3.88 \times 10^{-2}$ & $1.10 \times 10^{0}$ & $1.09 \times 10^{0}$  \\\hline
$\alpha_1 = 0,\alpha_2 = 1$ & $1.16 \times 10^{0}$ & $1.15 \times 10^{0}$ & $1.04 \times 10^{-2}$ & $7.27 \times 10^{-2}$  \\\hline
\end{tabular}
\end{table}
}
%\newpage

\subsection{Type-b system}
\begin{figure}
     \centering
     \begin{subfigure}[b]{0.45\textwidth}
         \centering
         \includegraphics[width=\textwidth]{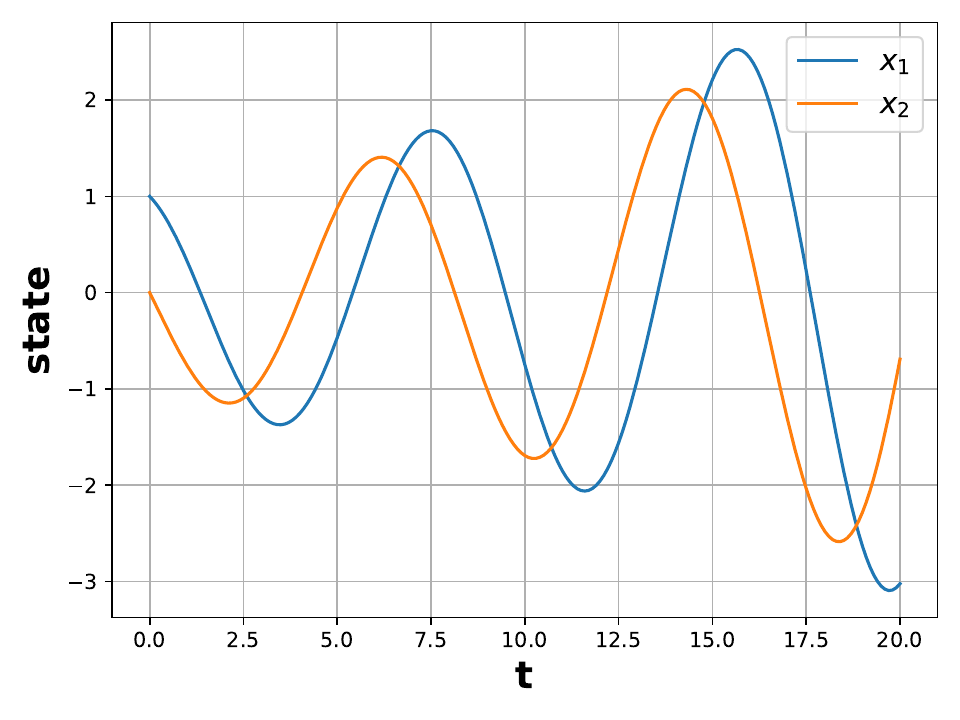}
         \caption{evolution of states}
         %\label{}
     \end{subfigure}
     \hfill
     \begin{subfigure}[b]{0.45\textwidth}
         \centering
         \includegraphics[width=\textwidth]{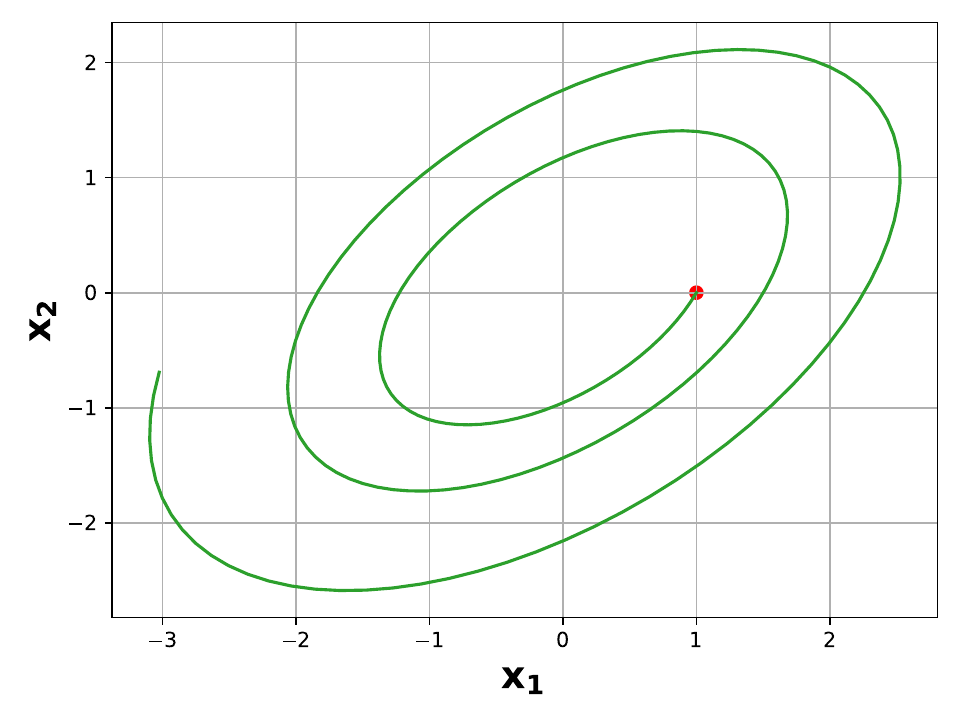}
         \caption{phase space}
         %\label{}
     \end{subfigure}
        \caption{Type-b system, $f_1 = x_2 - 0.4 x_1, \  f_2 = -0.8 x_1 + 0.5 x_2$}
        \label{type2_sol}
\end{figure}

{\label{Type_b_func_section}}
We consider Type-b system with

\begin{equation}
     \begin{split}
         & f_1 = x_2 - 0.4 x_1   \\
         & f_2 = -0.8 x_1 + 0.5 x_2
     \end{split}
 \end{equation}
 
with initial condition as $x_1(0) = 1 , x_2(0) =0 $. The system is solved upto t = 20 unit. The nature of solution is shown in figure \ref{type2_sol} 

We train deep hidden physics model using 201 data points from reference solution of states $x_1, x_2$ and additionally we consider 1000 random collocation points. The network architecture for $\mathscr{N}_s$ consists of 2 hidden layers with 64 neurons each and network $\mathscr{N}_f$ consists of 2 hidden layers with 64 neurons each. We train the model for 25000 epochs with learning rate of $10^{-3}$. The model is trained by considering states contributing to net data loss by choosing values of $\alpha_1, \alpha_2$.

\subsubsection{\label{type2_a1_1_a2_1} Case1: $\alpha_1 = 1,\alpha_2 = 1$ }
\begin{figure}
     \centering
     \begin{subfigure}[b]{0.45\textwidth}
         \centering
         \includegraphics[width=.95\linewidth]{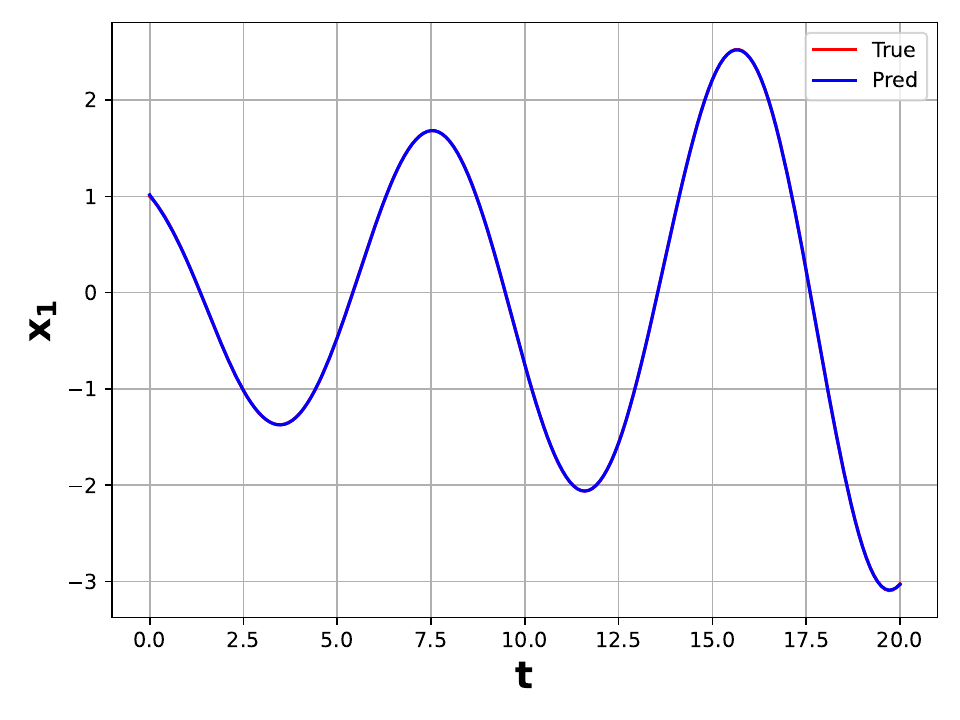}
         \caption{reference Vs predicted of state $x_1$ }
         %\label{}
     \end{subfigure}
     %\hfill
     \begin{subfigure}[b]{0.45\textwidth}
         \centering
         \includegraphics[width=.95\linewidth]{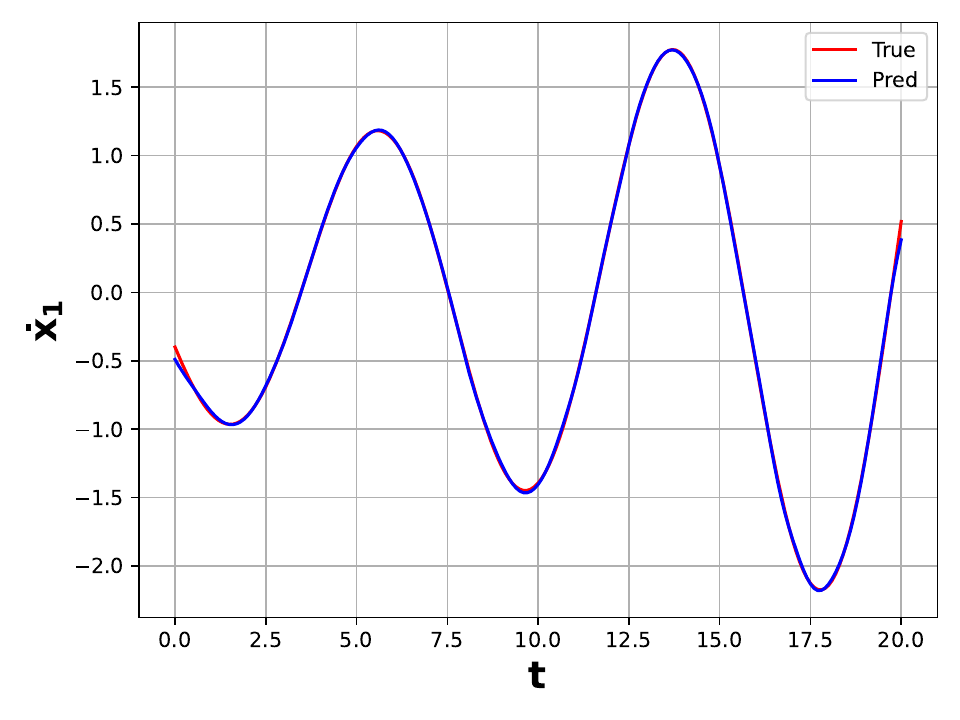}
         \caption{reference Vs predicted dynamics $\dot{x}_1$}
         %\label{}
     \end{subfigure}
     %\hfill
     \begin{subfigure}[b]{0.45\textwidth}
         \centering
         \includegraphics[width=.95\linewidth]{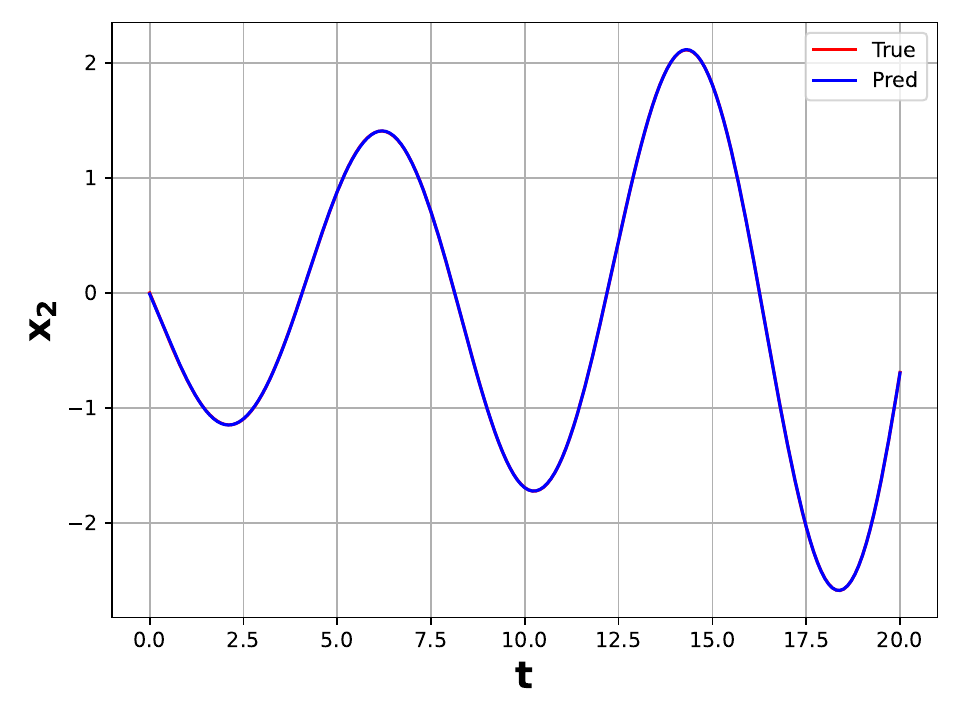}
         \caption{reference Vs predicted of state $x_2$}
         %\label{}
     \end{subfigure}
     %\hfill
     \begin{subfigure}[b]{0.45\textwidth}
         \centering
         \includegraphics[width=.95\linewidth]{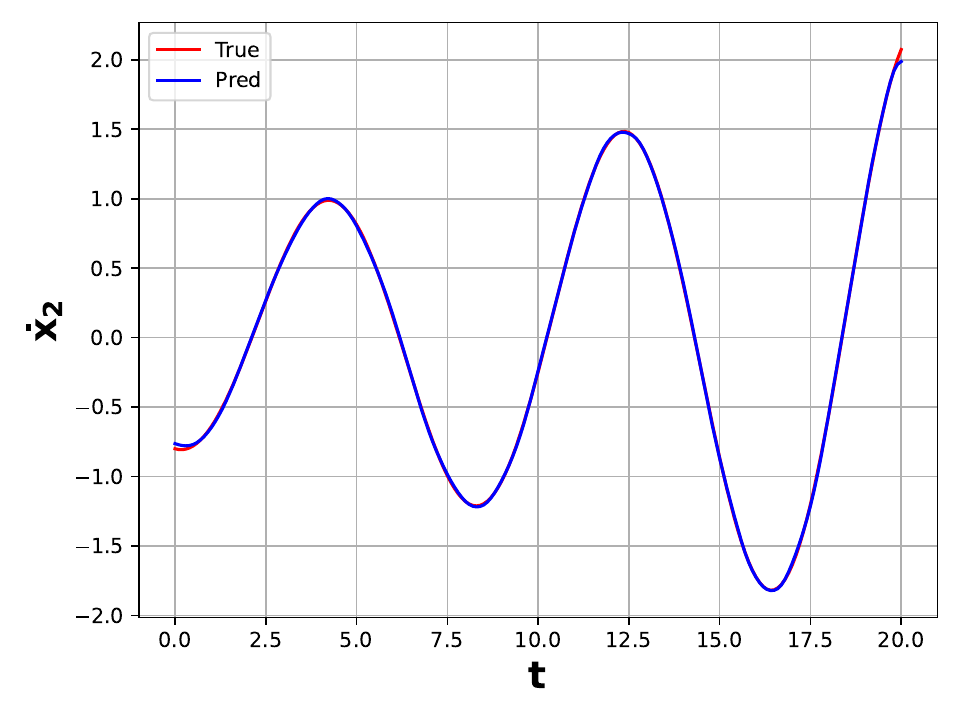}
         \caption{reference Vs predicted dynamics $\dot{x}_2$}
         %\label{}
     \end{subfigure}
        \caption{Type-b system, training with $\alpha_1 = 1,\alpha_2 = 1$ }
        \label{type2_a1_1_a2_1_sol}
\end{figure}

This shows that the data loss of both the states are contributing to net data loss. 
The results are shown in figure  \ref{type2_a1_1_a2_1_sol}

\subsubsection{\label{type2_a1_1_a2_0} Case2: $\alpha_1 = 1,\alpha_2 = 0$ }
\begin{figure}
     \centering
     \begin{subfigure}[b]{0.45\textwidth}
         \centering
         \includegraphics[width=.95\linewidth]{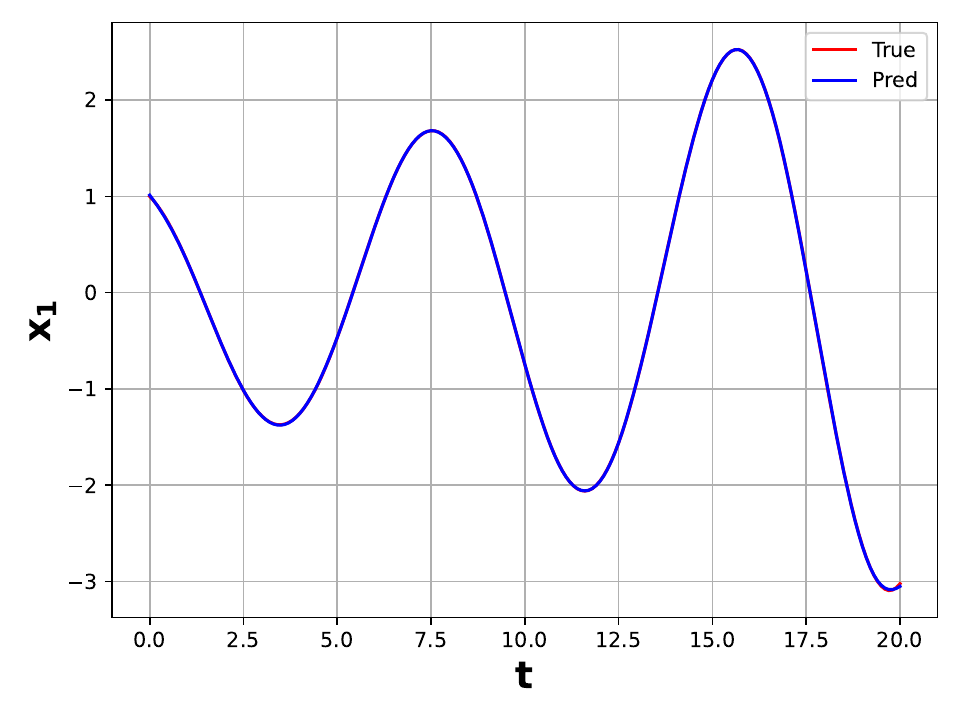}
         \caption{reference Vs predicted of state $x_1$ }
         %\label{}
     \end{subfigure}
     %\hfill
     \begin{subfigure}[b]{0.45\textwidth}
         \centering
         \includegraphics[width=.95\linewidth]{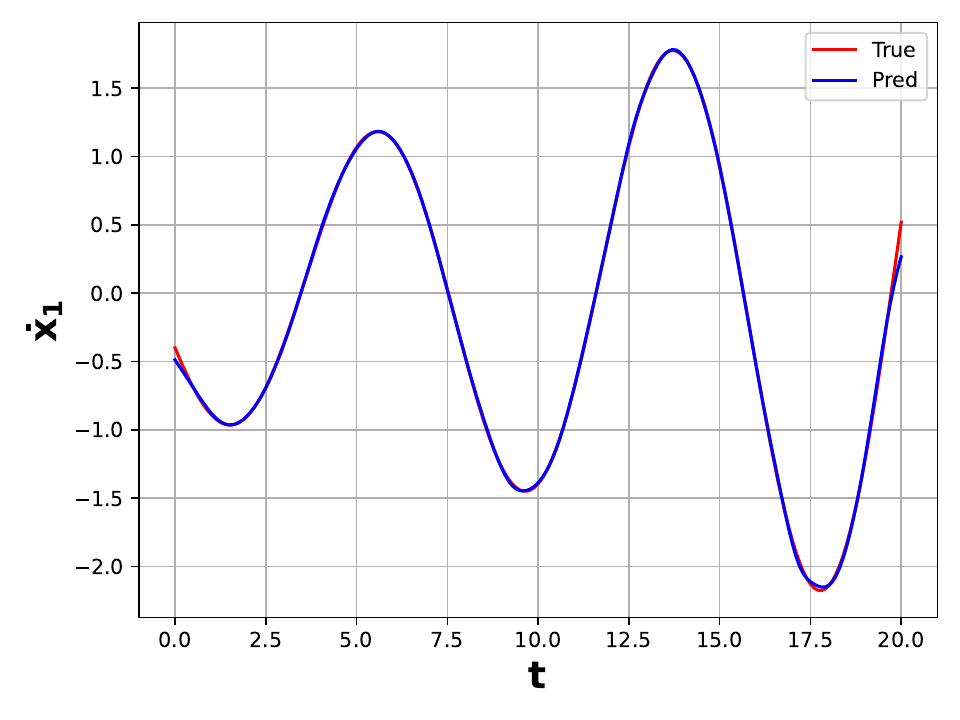}
         \caption{reference Vs predicted dynamics $\dot{x}_1$}
         %\label{}
     \end{subfigure}
     %\hfill
     \begin{subfigure}[b]{0.45\textwidth}
         \centering
         \includegraphics[width=.95\linewidth]{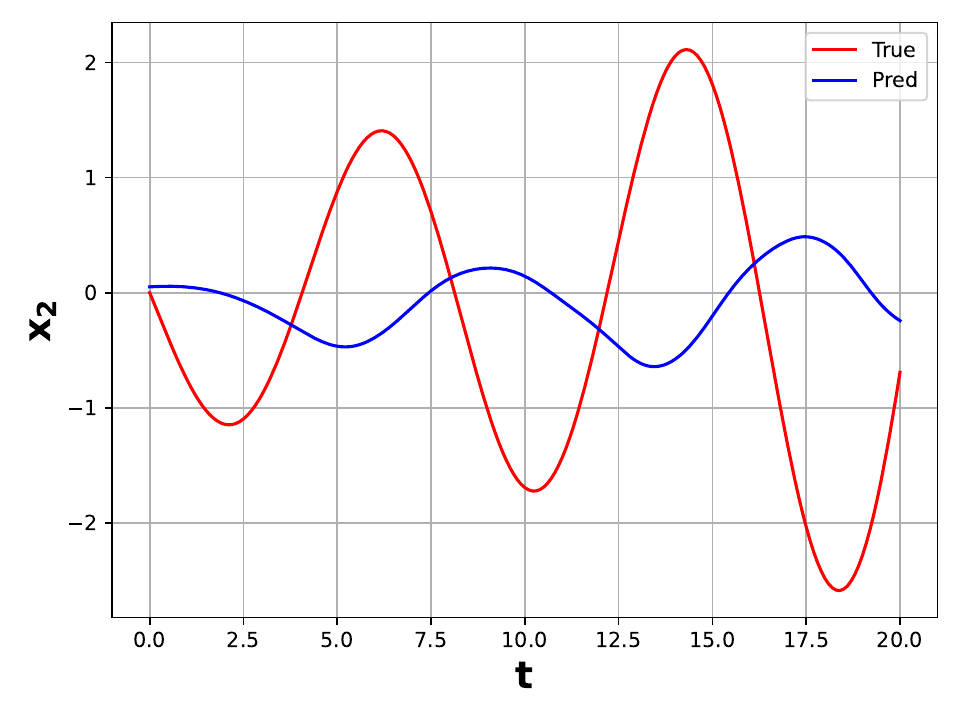}
         \caption{reference Vs predicted of state $x_2$}
         %\label{}
     \end{subfigure}
     %\hfill
     \begin{subfigure}[b]{0.45\textwidth}
         \centering
         \includegraphics[width=.95\linewidth]{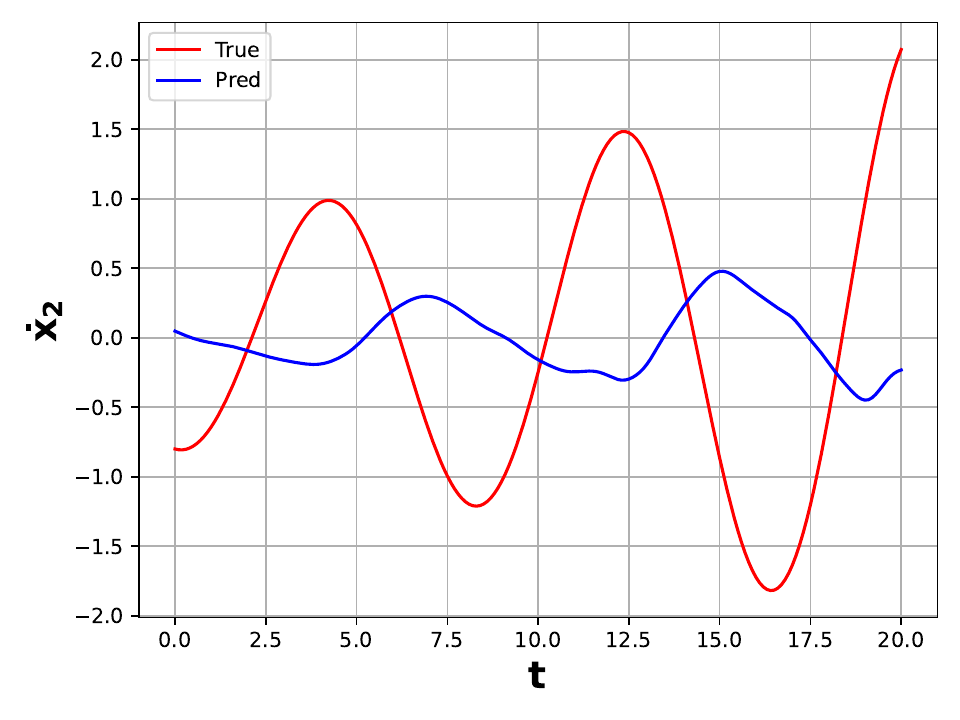}
         \caption{reference Vs predicted dynamics $\dot{x}_2$}
         %\label{}
     \end{subfigure}
        \caption{Type-b system, training with $\alpha_1 = 1,\alpha_2 = 0$  }
        \label{type2_a1_1_a2_0_sol}
\end{figure}

This shows that the data loss of only state $x_1$ is contributing to the net data loss. The results are shown in figure  \ref{type2_a1_1_a2_0_sol}

\subsubsection{\label{type2_a1_0_a2_1} Case3: $\alpha_1 = 0,\alpha_2 = 1$ }
\begin{figure}
     \centering
     \begin{subfigure}[b]{0.45\textwidth}
         \centering
         \includegraphics[width=.95\linewidth]{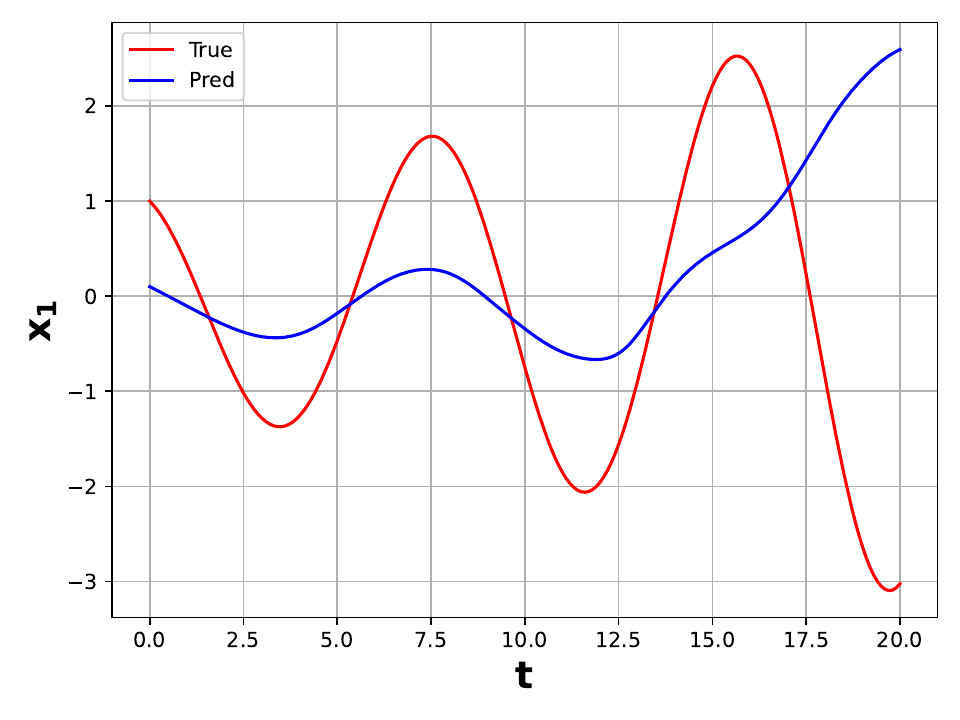}
         \caption{reference Vs predicted of state $x_1$ }
         %\label{}
     \end{subfigure}
     %\hfill
     \begin{subfigure}[b]{0.45\textwidth}
         \centering
         \includegraphics[width=.95\linewidth]{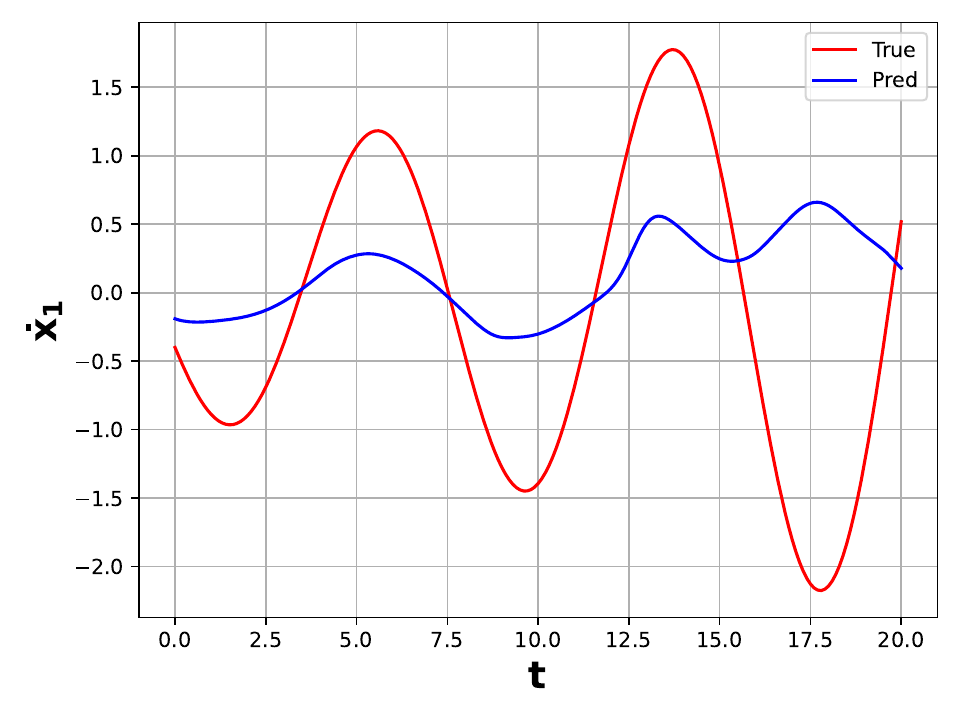}
         \caption{reference Vs predicted dynamics $\dot{x}_1$}
         %\label{}
     \end{subfigure}
     %\hfill
     \begin{subfigure}[b]{0.45\textwidth}
         \centering
         \includegraphics[width=.95\linewidth]{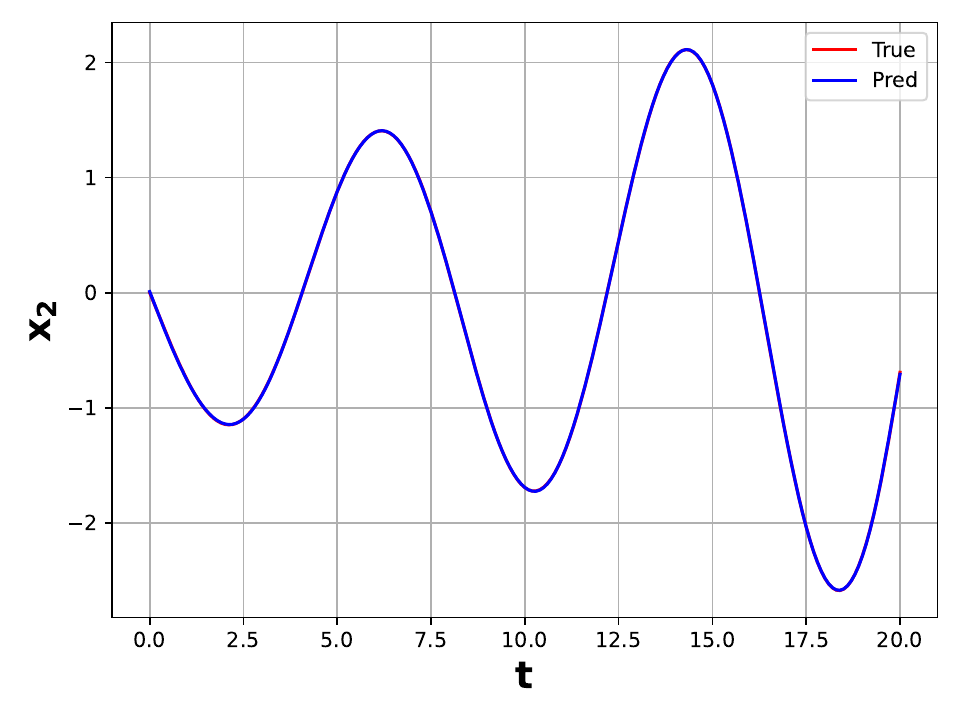}
         \caption{reference Vs predicted of state $x_2$}
         %\label{}
     \end{subfigure}
     %\hfill
     \begin{subfigure}[b]{0.45\textwidth}
         \centering
         \includegraphics[width=.95\linewidth]{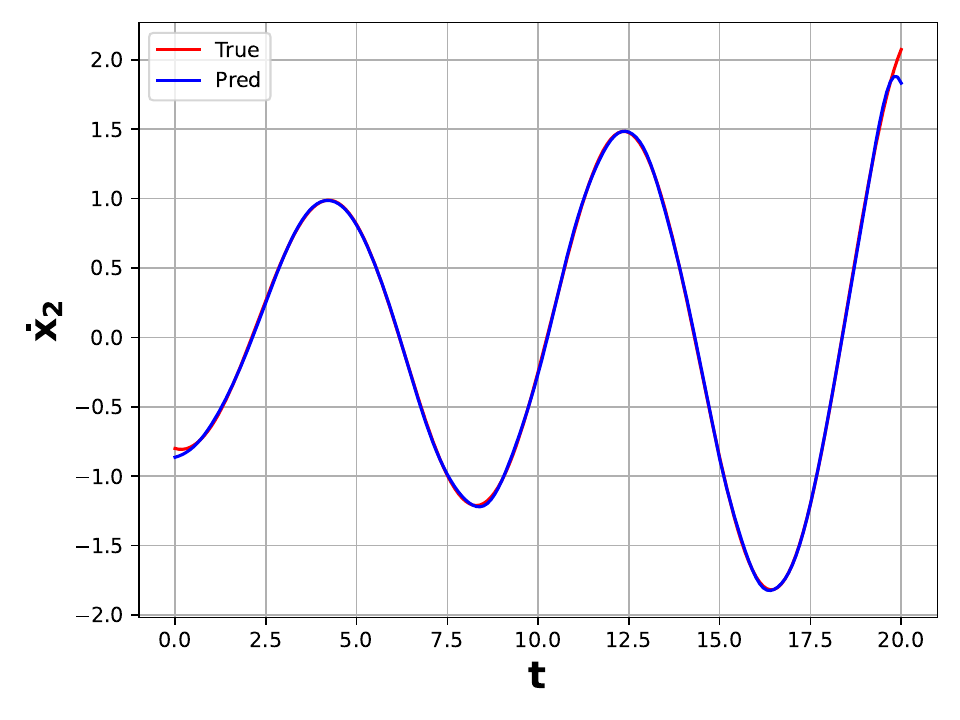}
         \caption{reference Vs predicted dynamics $\dot{x}_2$}
         %\label{}
     \end{subfigure}
        \caption{Type-b system, training with $\alpha_1 = 0,\alpha_2 = 1$}
        \label{type2_a1_0_a2_1_sol}
\end{figure}

This shows that the data loss of only state $x_2$ is contributing to the net data loss. The results are shown in figure  \ref{type2_a1_0_a2_1_sol}

{\renewcommand{\arraystretch}{1.2}
\begin{table}
\caption{$L_2$ Error for Type-b system with different contributions in data loss}
\label{type_b_error}
\begin{tabular}{|l||*{4}{c|}}\hline
%\backslashbox{$\alpha$}{quantities}
&\makebox[5.5em]{$x_1$}&\makebox[5.5em]{$\dot{x}_1$}&\makebox[5.5em]{$x_2$}
&\makebox[5.5em]{$\dot{x}_2$}\\\hline
$\alpha_1 = 1,\alpha_2 = 1$ & $1.78 \times 10^{-3}$ & $1.52 \times 10^{-2}$  & $1.55 \times 10^{-3}$  & $1.07\times 10^{-2}$  \\\hline
$\alpha_1 = 1,\alpha_2 = 0$ & $2.43 \times 10^{-3}$ & $2.33 \times 10^{-2}$  & $1.16 \times 10^{0}$  & $1.17 \times 10^{0}$  \\\hline
$\alpha_1 = 0,\alpha_2 = 1$ & $1.21 \times 10^{0}$ & $1.00 \times 10^{0}$  & $2.27 \times 10^{-3}$  & $2.30 \times 10^{-2}$  \\\hline
\end{tabular}
\end{table}
}
%\newpage

\subsection{Nonlinear system}
\begin{figure}
     \centering
     \begin{subfigure}[b]{0.45\textwidth}
         \centering
         \includegraphics[width=\textwidth]{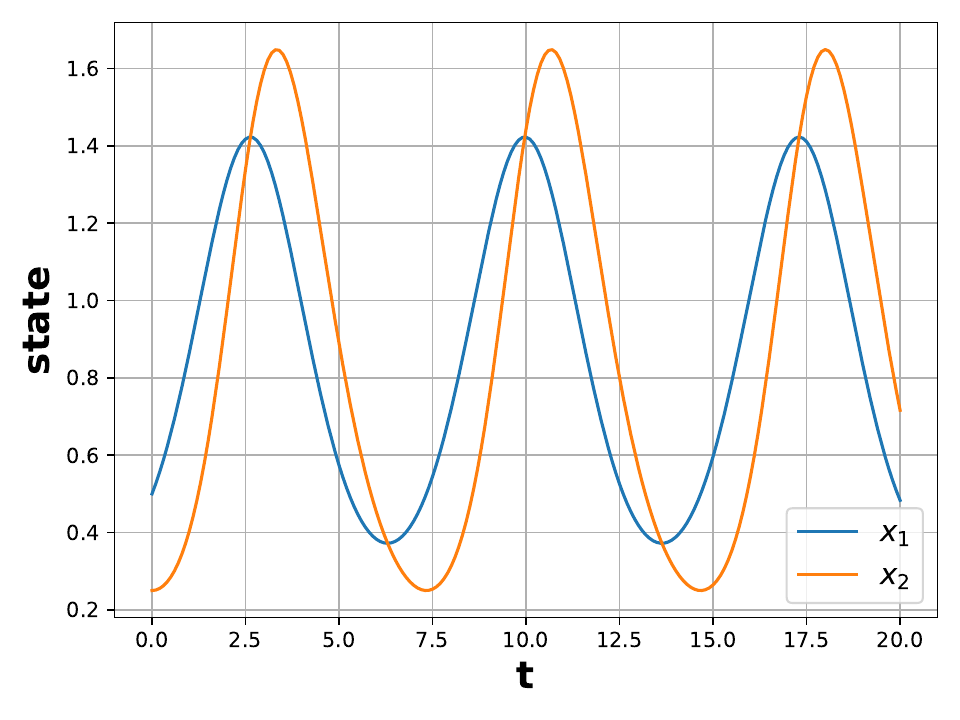}
         \caption{evolution of states}
         %\label{}
     \end{subfigure}
     \hfill
     \begin{subfigure}[b]{0.45\textwidth}
         \centering
         \includegraphics[width=\textwidth]{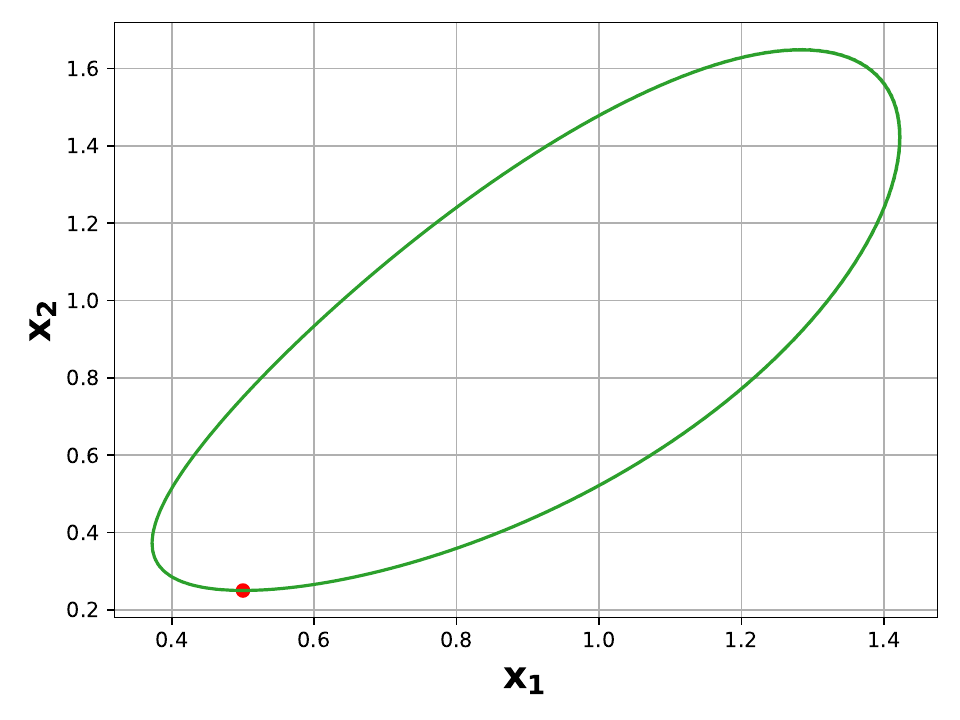}
         \caption{phase space}
         %\label{}
     \end{subfigure}
        \caption{Nonlinear system, $f_1 = x_1 - x_2, \  f_2 = x_1^2- x_2$}
        \label{nonlinear_sol}
\end{figure}

{\label{nonlinear_func_section}}
Let's consider non-linear system given by,

\begin{equation}
     \begin{split}
         & f_1 = x_1 - x_2   \\
         & f_2 = x_1^2- x_2
     \end{split}
 \end{equation}

with initial condition as $x_1(0) = 0.5 , x_2(0) = 0.25 $. The system is solved upto t = 20 unit. The nature of solution is shown in figure \ref{nonlinear_sol} 

Next, We train deep hidden physics model using 201 data points from reference solution of states $x_1, x_2$ of nonlinear system and additionally we consider 1000 random collocation points. The network architecture for $\mathscr{N}_s$ consists of 2 hidden layers with 64 neurons each and network $\mathscr{N}_f$ consists of 2 hidden layers with 64 neurons each. We train the model for 25000 epochs with learning rate of $10^{-3}$. The model is trained by considering states contributing to net data loss by choosing values of $\alpha_1, \alpha_2$.

\subsubsection{\label{nonlinear_a1_1_a2_1} Case1: $\alpha_1 = 1,\alpha_2 = 1$ }
\begin{figure}
     \centering
     \begin{subfigure}[b]{0.45\textwidth}
         \centering
         \includegraphics[width=.95\linewidth]{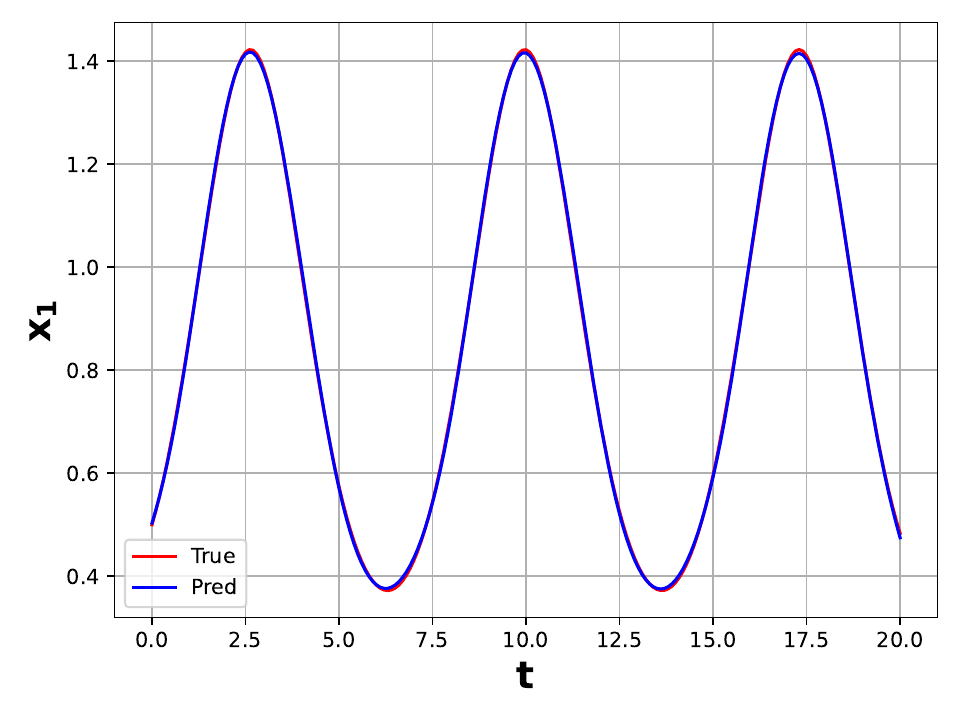}
         \caption{reference Vs predicted of state $x_1$ }
         %\label{}
     \end{subfigure}
     %\hfill
     \begin{subfigure}[b]{0.45\textwidth}
         \centering
         \includegraphics[width=.95\linewidth]{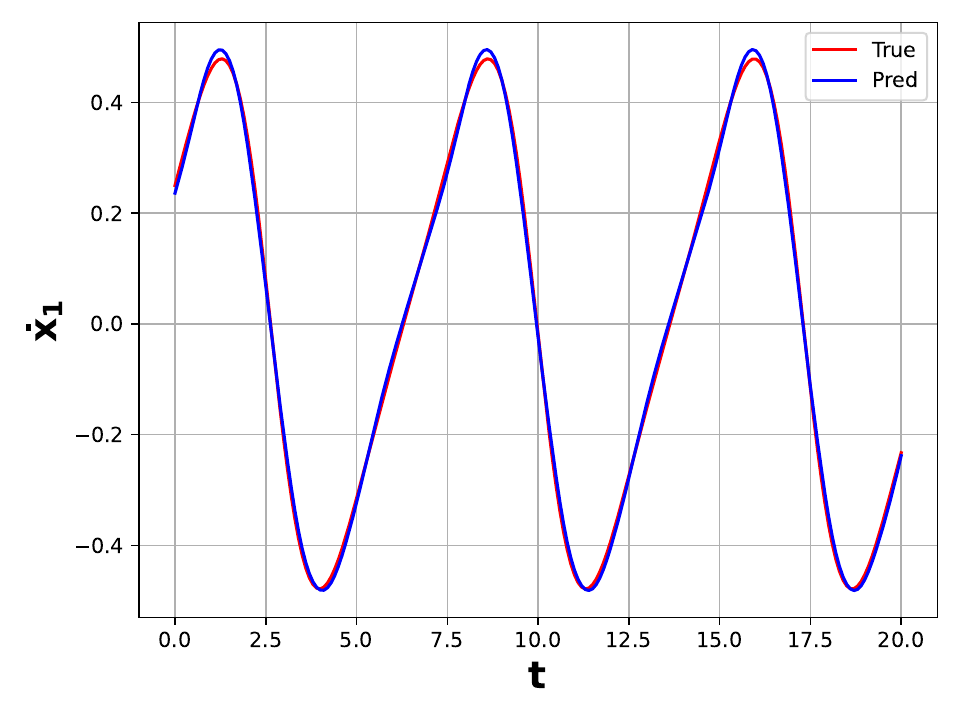}
         \caption{reference Vs predicted dynamics $\dot{x}_1$}
         %\label{}
     \end{subfigure}
     %\hfill
     \begin{subfigure}[b]{0.45\textwidth}
         \centering
         \includegraphics[width=.95\linewidth]{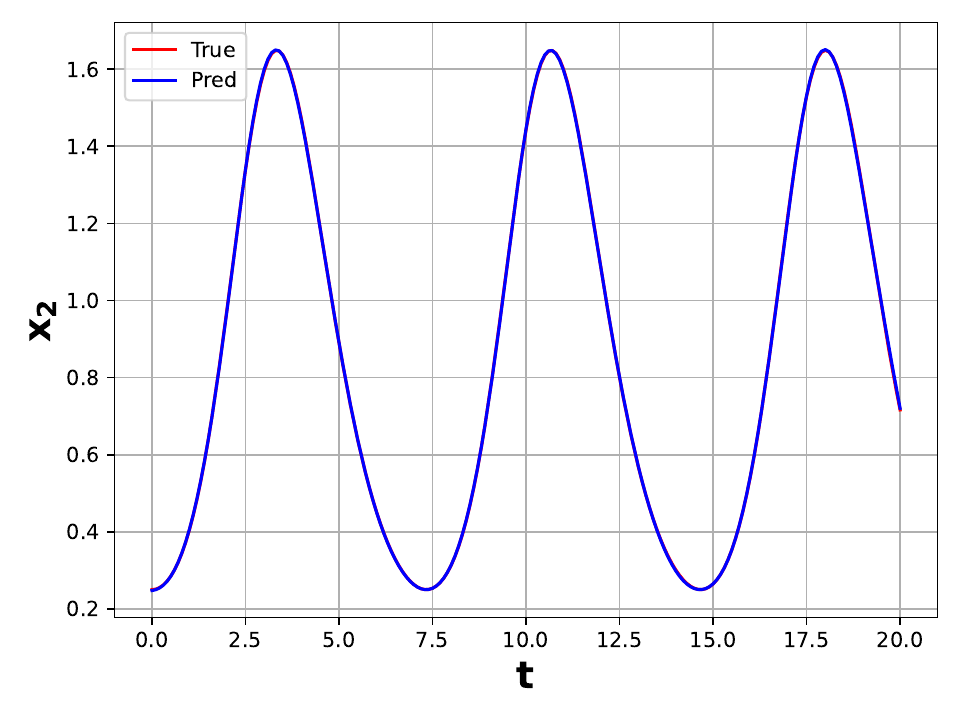}
         \caption{reference Vs predicted of state $x_2$}
         %\label{}
     \end{subfigure}
     %\hfill
     \begin{subfigure}[b]{0.45\textwidth}
         \centering
         \includegraphics[width=.95\linewidth]{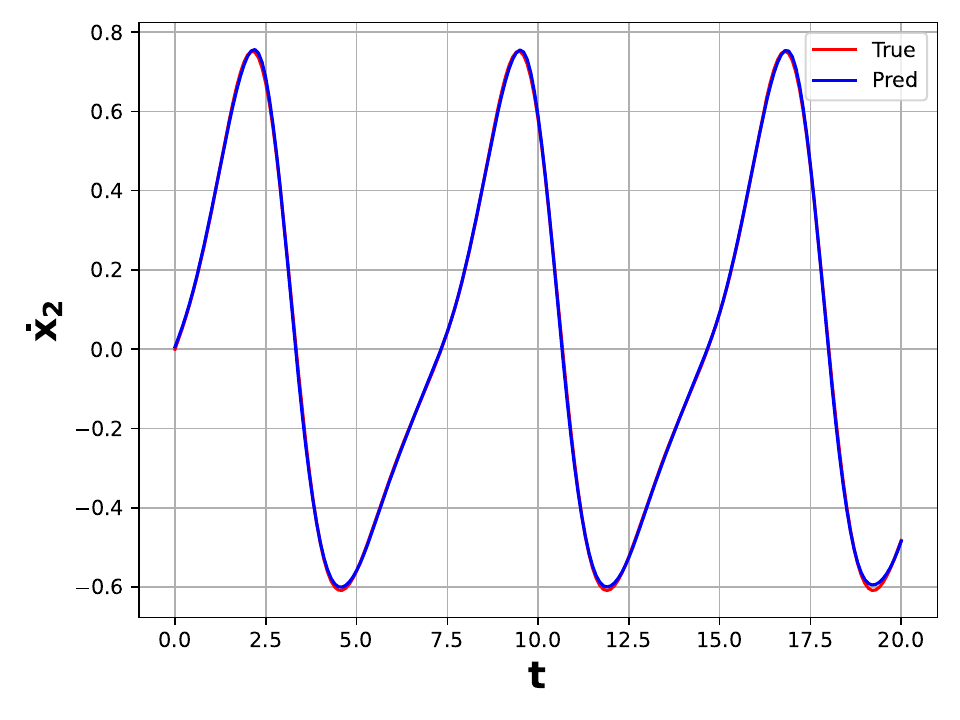}
         \caption{reference Vs predicted dynamics $\dot{x}_2$}
         %\label{}
     \end{subfigure}
        \caption{Nonlinear system, training with $\alpha_1 = 1,\alpha_2 = 1$ }
        \label{nonlinear_a1_1_a2_1_sol}
\end{figure}

This shows that the data loss of both the states are contributing to net data loss. 
The results are shown in figure  \ref{nonlinear_a1_1_a2_1_sol}

\subsubsection{\label{nonlinear_a1_1_a2_0} Case2: $\alpha_1 = 1,\alpha_2 = 0$ }
\begin{figure}
     \centering
     \begin{subfigure}[b]{0.45\textwidth}
         \centering
         \includegraphics[width=.95\linewidth]{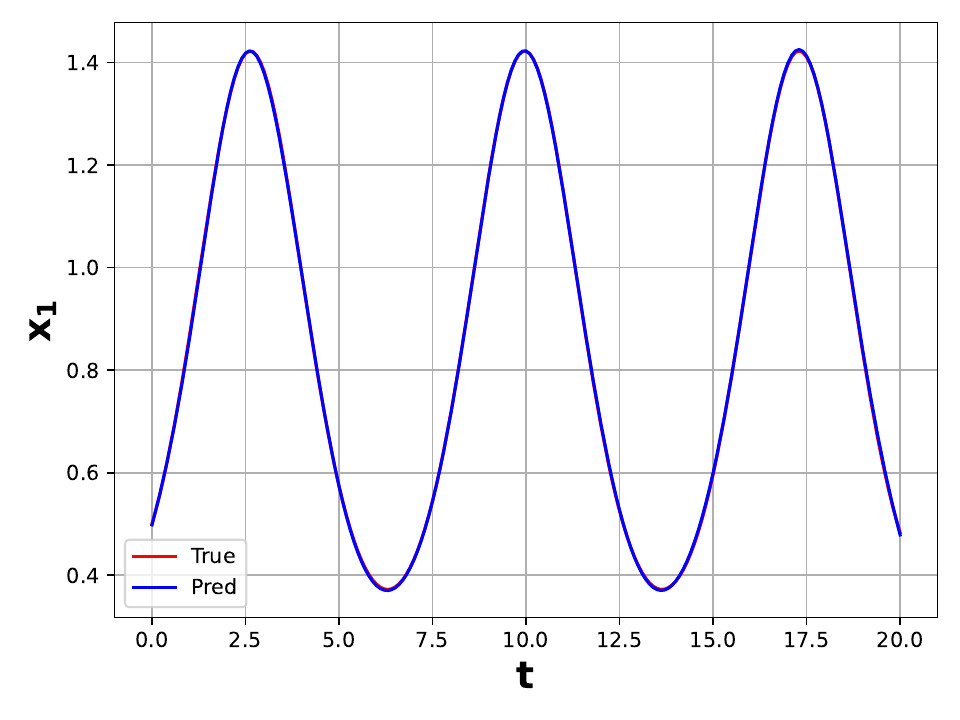}
         \caption{reference Vs predicted of state $x_1$ }
         %\label{}
     \end{subfigure}
     %\hfill
     \begin{subfigure}[b]{0.45\textwidth}
         \centering
         \includegraphics[width=.95\linewidth]{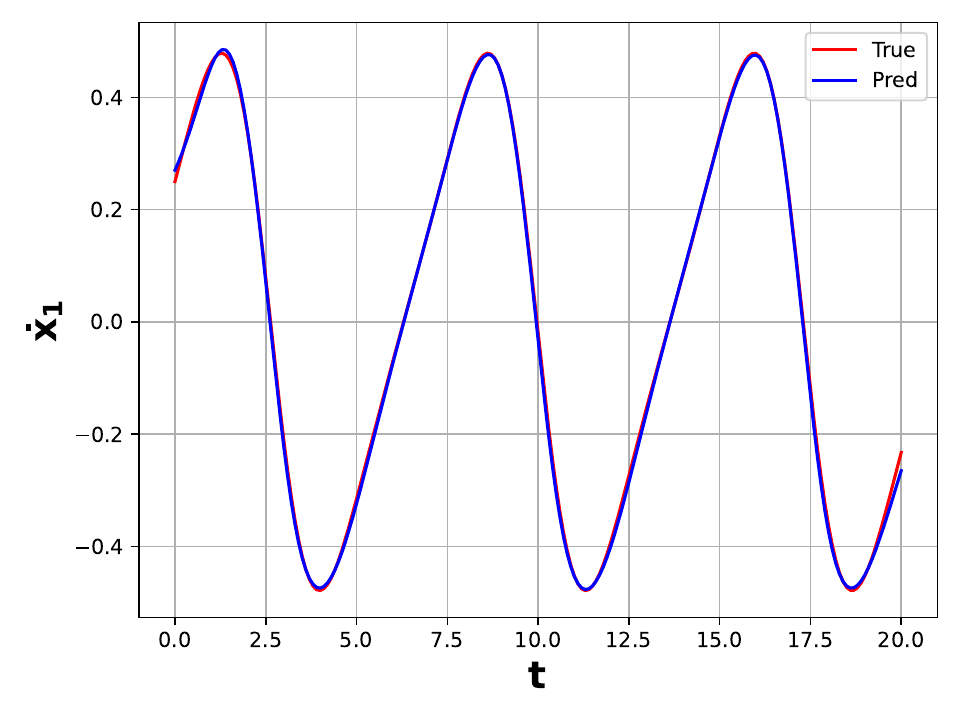}
         \caption{reference Vs predicted dynamics $\dot{x}_1$}
         %\label{}
     \end{subfigure}
     %\hfill
     \begin{subfigure}[b]{0.45\textwidth}
         \centering
         \includegraphics[width=.95\linewidth]{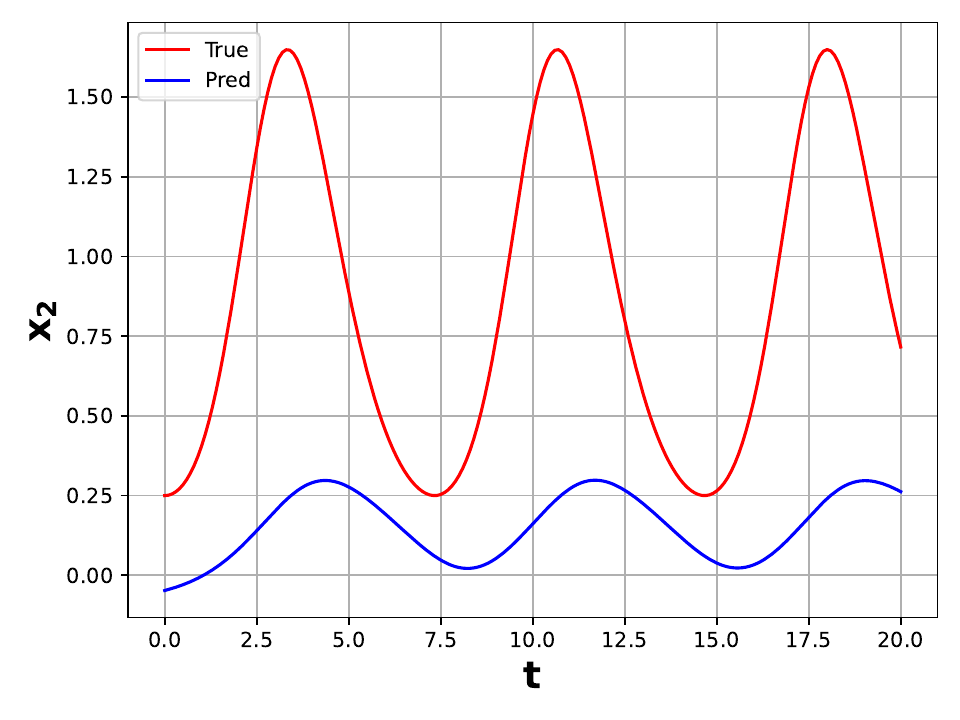}
         \caption{reference Vs predicted of state $x_2$}
         %\label{}
     \end{subfigure}
     %\hfill
     \begin{subfigure}[b]{0.45\textwidth}
         \centering
         \includegraphics[width=.95\linewidth]{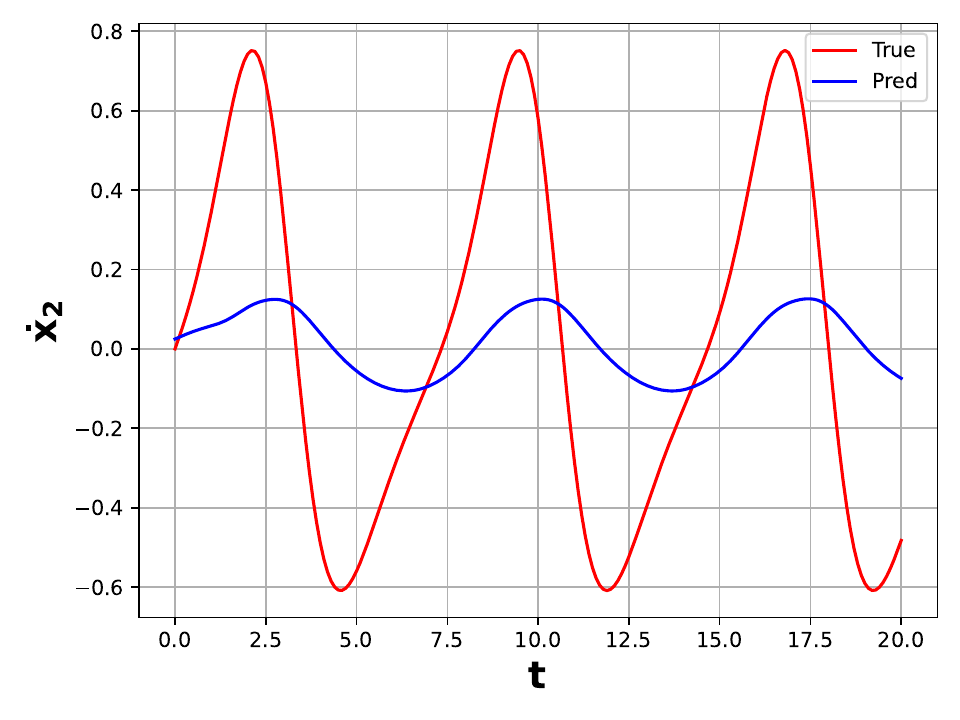}
         \caption{reference Vs predicted dynamics $\dot{x}_2$}
         %\label{}
     \end{subfigure}
        \caption{Nonlinear system, training with $\alpha_1 = 1,\alpha_2 = 0$  }
        \label{nonlinear_a1_1_a2_0_sol}
\end{figure}

This shows that the data loss of only state $x_1$ is contributing to the net data loss. The results are shown in figure  \ref{nonlinear_a1_1_a2_0_sol}

\subsubsection{\label{nonlinear_a1_0_a2_1} Case3: $\alpha_1 = 0,\alpha_2 = 1$ }
\begin{figure}
     \centering
     \begin{subfigure}[b]{0.45\textwidth}
         \centering
         \includegraphics[width=.95\linewidth]{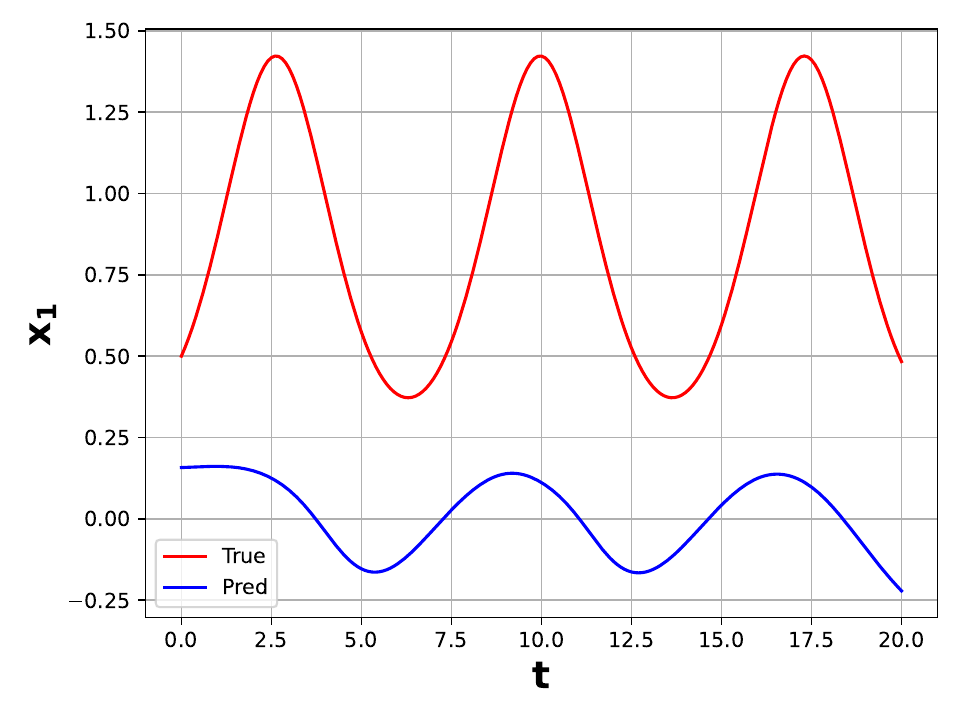}
         \caption{reference Vs predicted of state $x_1$ }
         %\label{}
     \end{subfigure}
     %\hfill
     \begin{subfigure}[b]{0.45\textwidth}
         \centering
         \includegraphics[width=.95\linewidth]{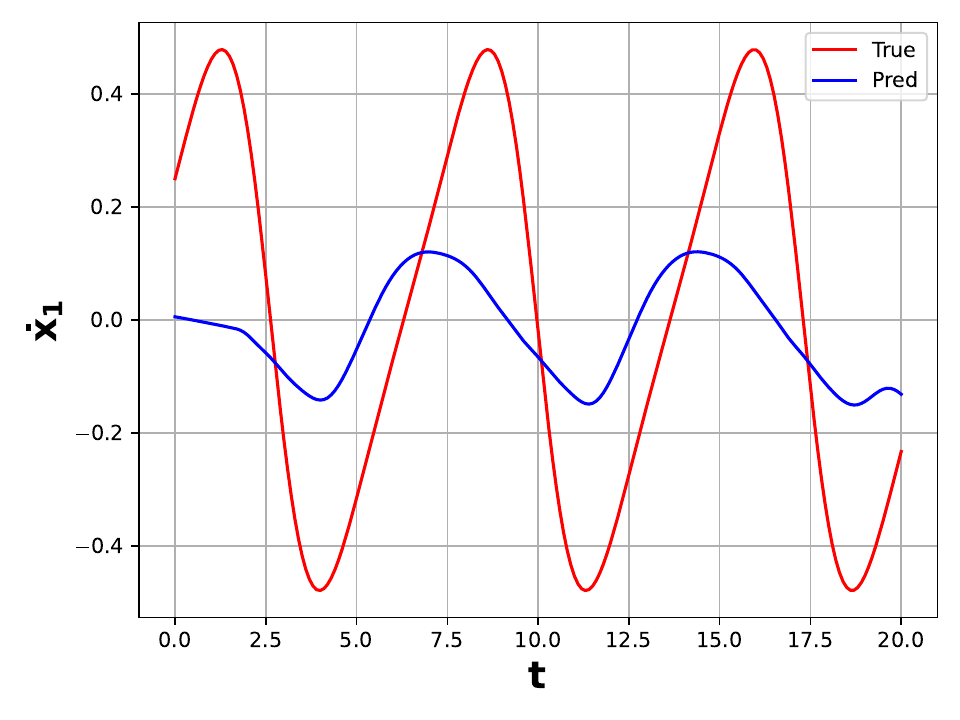}
         \caption{reference Vs predicted dynamics $\dot{x}_1$}
         %\label{}
     \end{subfigure}
     %\hfill
     \begin{subfigure}[b]{0.45\textwidth}
         \centering
         \includegraphics[width=.95\linewidth]{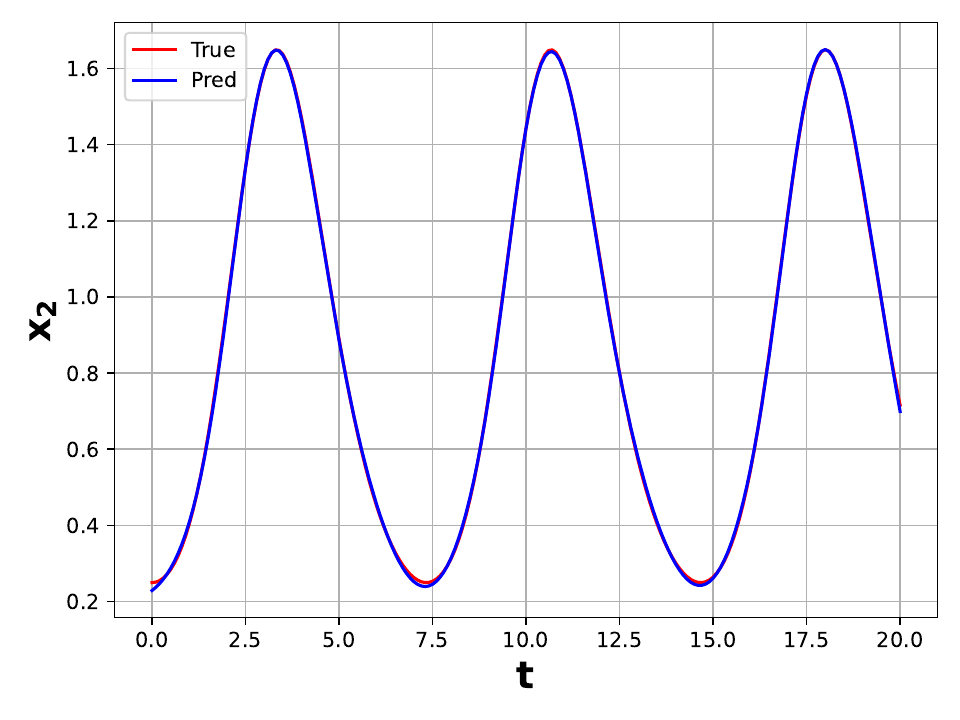}
         \caption{reference Vs predicted of state $x_2$}
         %\label{}
     \end{subfigure}
     %\hfill
     \begin{subfigure}[b]{0.45\textwidth}
         \centering
         \includegraphics[width=.95\linewidth]{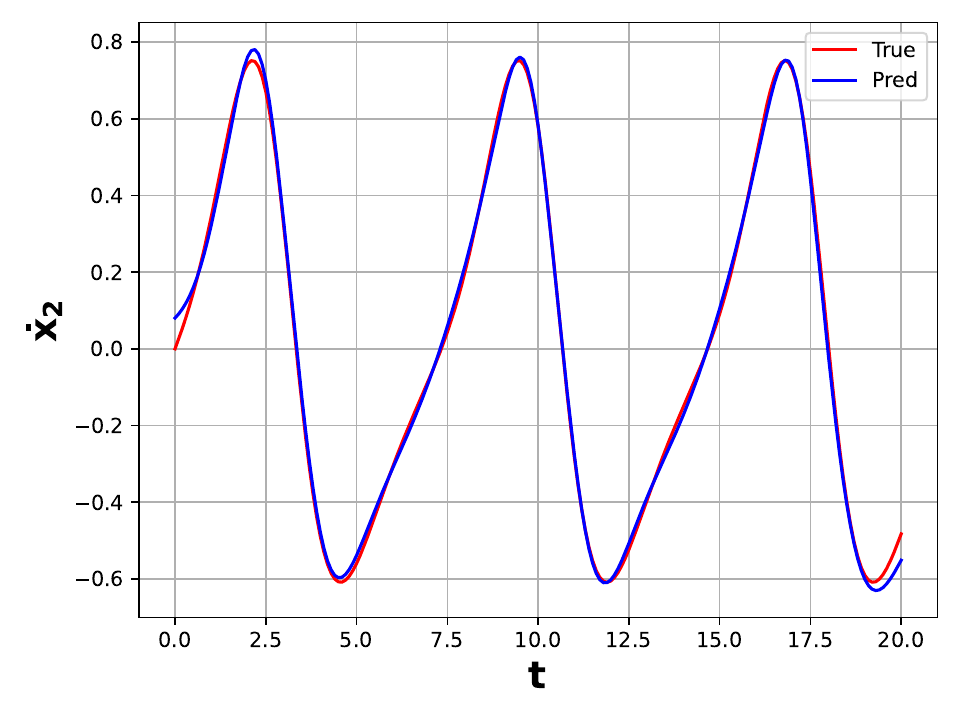}
         \caption{reference Vs predicted dynamics $\dot{x}_2$}
         %\label{}
     \end{subfigure}
        \caption{Nonlinear system, training with $\alpha_1 = 0,\alpha_2 = 1$}
        \label{nonlinear_a1_0_a2_1_sol}
\end{figure}

This shows that the data loss of only state $x_2$ is contributing to the net data loss. The results are shown in figure  \ref{nonlinear_a1_0_a2_1_sol}

{\renewcommand{\arraystretch}{1.2}
\begin{table}
\caption{$L_2$ Error for Nonlinear system with different contributions in data loss}
\label{nonlinear_error}
\begin{tabular}{|l||*{4}{c|}}\hline
%\backslashbox{$\alpha$}{quantities}
&\makebox[5.5em]{$x_1$}&\makebox[5.5em]{$\dot{x}_1$}&\makebox[5.5em]{$x_2$}
&\makebox[5.5em]{$\dot{x}_2$}\\\hline 
$\alpha_1 = 1,\alpha_2 = 1$ & $4.44 \times 10^{-3}$ & $3.28 \times 10^{-2}$  & $2.25 \times 10^{-3}$  & $1.36\times 10^{-2}$  \\\hline
$\alpha_1 = 1,\alpha_2 = 0$ & $2.51 \times 10^{-3}$ & $2.34 \times 10^{-2}$  & $8.35 \times 10^{-1}$  & $8.96 \times 10^{-1}$  \\\hline
$\alpha_1 = 0,\alpha_2 = 1$ & $9.64 \times 10^{-1}$ & $8.29 \times 10^{-1}$  & $5.20 \times 10^{-3}$  & $4.21 \times 10^{-2}$  \\\hline
\end{tabular}
\end{table}
}
%\newpage
As we can observe in Table \ref{type_a_error}, \ref{type_b_error} and \ref{nonlinear_error}, When data losses of both the states are contributing to the total losses i.e. $\alpha_1 = 1,\alpha_2 = 1$ . The $L_2$ Error of states $x_1,x_2$ are of order $\mathcal{O}(10^{-3})$ and of dynamics  $\dot{x}_1, \dot{x}_2$ are of order $\mathcal{O}(10^{-2})$, that means the model is able to perfectly capture the state and dynamics altogether.
However in other cases, If the data loss of only one of the state is given, then the error for that state and dynamics are of order $\mathcal{O}(10^{-2})$. Its interesting to note that even though the other state information is not given to model, it's perfectly able to predict dynamics of given data state. The state of which data is not included in the training, the errors are of order $\mathcal{O}(10^{0})$, which means it fails to capture the state and dynamics both.

\subsection{Lorenz system}
\begin{figure}
     \centering
     \begin{subfigure}[b]{0.45\textwidth}
         \centering
         \includegraphics[width=\textwidth]{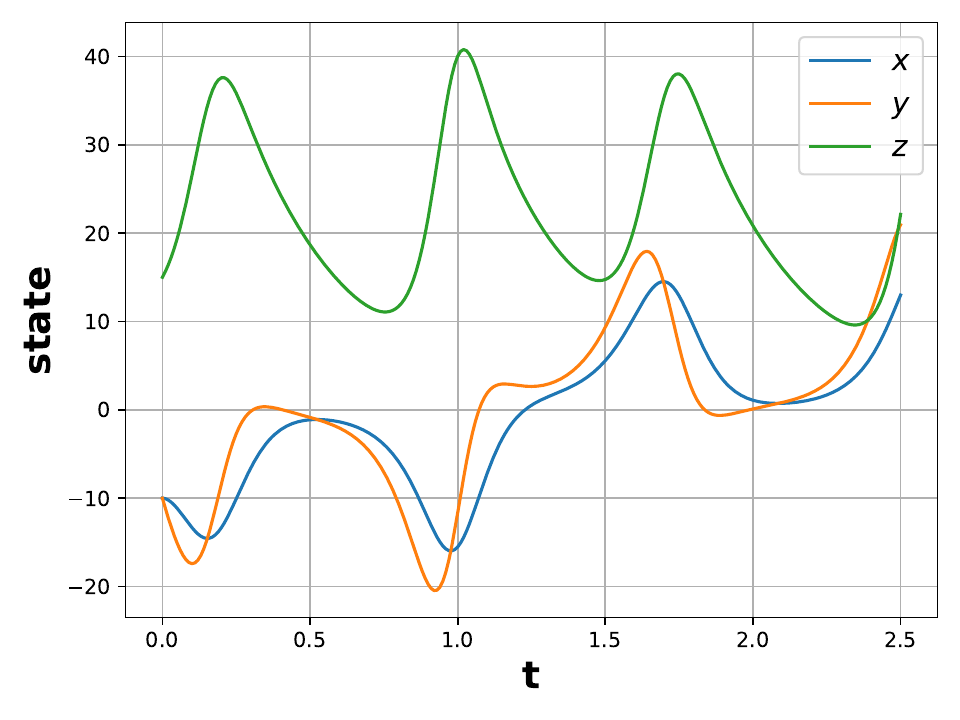}
         \caption{evolution of states}
         %\label{}
     \end{subfigure}
     \hfill
     \begin{subfigure}[b]{0.5\textwidth}
         \centering
         \includegraphics[width=\textwidth]{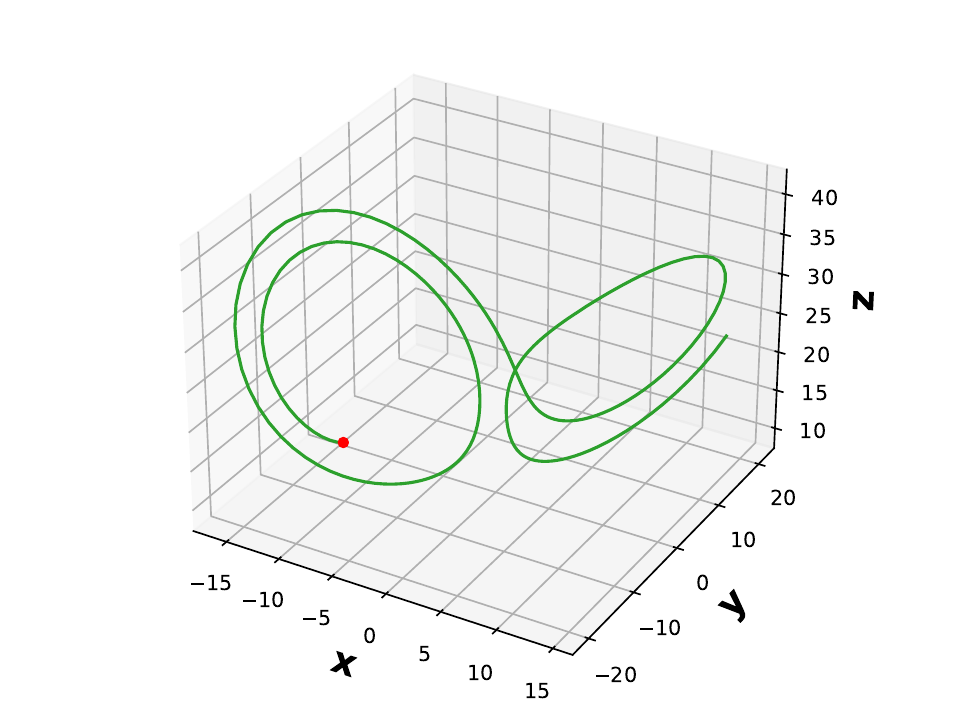}
         \caption{phase space}
         %\label{}
     \end{subfigure}
        \caption{Lorenz system}
        \label{lorenz_sol}
\end{figure}

{\label{Type_lorentz_section}}
In chaos theory, the Lorenz system of equation is given by three ordinary differential equations as shown,

\begin{equation}
     \begin{split}
         &\frac{d x}{dt} = \sigma (y - x) \\
         &\frac{d y}{dt} = x (\rho - z) - y \\
         &\frac{d z}{dt} = xy - \beta z  
     \end{split}
 \end{equation}

where the parameters $\sigma$, $\rho$,  $\beta$ are positive, and $\sigma=10$, $\rho=28$ and $\beta=8/3$ results in chaotic solution \cite{hirsch2004differential}. We initialize the system at  $x(0)=-10$, $y(0)=-10$ and 
$z(0)=15$ and solve upto $t = 2.5$ unit. The solution is shown in the figure \ref{lorenz_sol}.

Here we train deep hidden physics model to discover the states and dynamics of lorentz system based on partial states solution.
We construct neural network 
$\mathscr{N}_s$ which takes input $[t]$ and outputs states $[\mathscr{N}_{x} ,\mathscr{N}_{y},\mathscr{N}_{z}]$ and other neural network $\mathscr{N}_f$ takes input $[\mathscr{N}_{x} ,\mathscr{N}_{y}, \mathscr{N}_{z},\mathscr{N}_{x}\mathscr{N}_{y},
\mathscr{N}_{y}\mathscr{N}_{z},\mathscr{N}_{z}\mathscr{N}_{x}]$ and outputs dynamics $[ \mathscr{N}_{\dot{x}},\mathscr{N}_{\dot{y}},\mathscr{N}_{\dot{z}}]$.

We construct the losses based on data and equations similar to as shown in section \ref{problem_setup} , First we define individual data losses as,

\begin{equation}
\mathcal{L}_{d,x} = \lVert x - \mathscr{N}_{x}\rVert_2,
\end{equation}

\begin{equation}
\mathcal{L}_{d,y} = \lVert y - \mathscr{N}_{y}\rVert_2,
\end{equation}

\begin{equation}
\mathcal{L}_{d,z} = \lVert z - \mathscr{N}_{z}\rVert_2,
\end{equation}

Next we define loss equation as,

\begin{equation}
\mathcal{L}_{eq,x} = \lVert \frac{d \mathscr{N}_{x}}{dt}  - \mathscr{N}_{\dot{x}} \rVert_2,
\end{equation}

\begin{equation}
\mathcal{L}_{eq,y} = \lVert \frac{d \mathscr{N}_{y}}{dt}  - \mathscr{N}_{\dot{y}} \rVert_2,
\end{equation}

\begin{equation}
\mathcal{L}_{eq,z} = \lVert \frac{d \mathscr{N}_{z}}{dt}  - \mathscr{N}_{\dot{z}} \rVert_2,
\end{equation}

We define net data loss as,
\begin{equation}
\mathcal{L}_{d} = \alpha_x \mathcal{L}_{d,x} + \alpha_y  \mathcal{L}_{d,y} + \alpha_z  \mathcal{L}_{d,z}
\end{equation}

We define net equation loss as,
\begin{equation}
\mathcal{L}_{eq} = \mathcal{L}_{eq,x} + \mathcal{L}_{eq,y} + \mathcal{L}_{eq,z}
\end{equation}

The total loss given by,

\begin{equation}
\mathcal{L}_{total} = \mathcal{L}_{d} + \mathcal{L}_{eq}
\end{equation}

The parameters of neural networks $\mathscr{N}_s$ ,  $\mathscr{N}_f$ are learned by minimizing the total loss $\mathcal{L}_{total}$. The model is trained by considering different contribution of states by changing  $\alpha_x, \alpha_y$ and  $\alpha_z$. We also non-dimensionalize the equations and data before passing it to the neural network, this is to make ensure that input and outputs variables are in a reasonable range, while prediction we re-scale back to original values.

\subsubsection{\label{type_LS_a1_1_a2_1_a3_1} Case1: $\alpha_x = 1,\alpha_y = 1, \alpha_z = 1$ }
\begin{figure}
     \centering
     \begin{subfigure}[b]{0.45\textwidth}
         \centering
         \includegraphics[width=.95\linewidth]{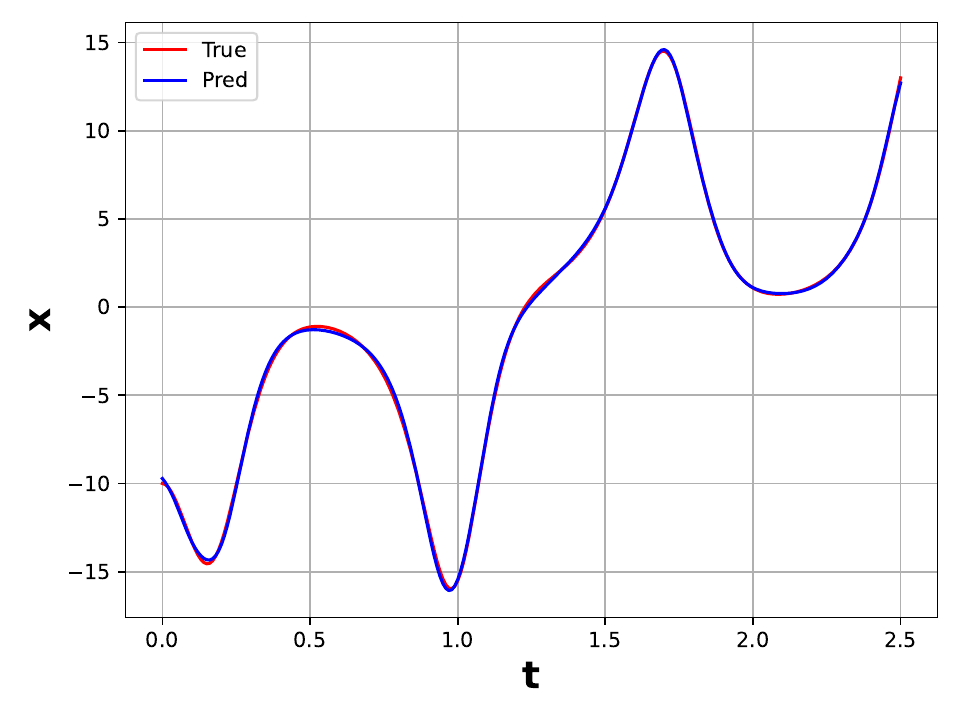}
         \caption{reference Vs predicted of state $x$ }
         %\label{}
     \end{subfigure}
     %\hfill
     \begin{subfigure}[b]{0.45\textwidth}
         \centering
         \includegraphics[width=.95\linewidth]{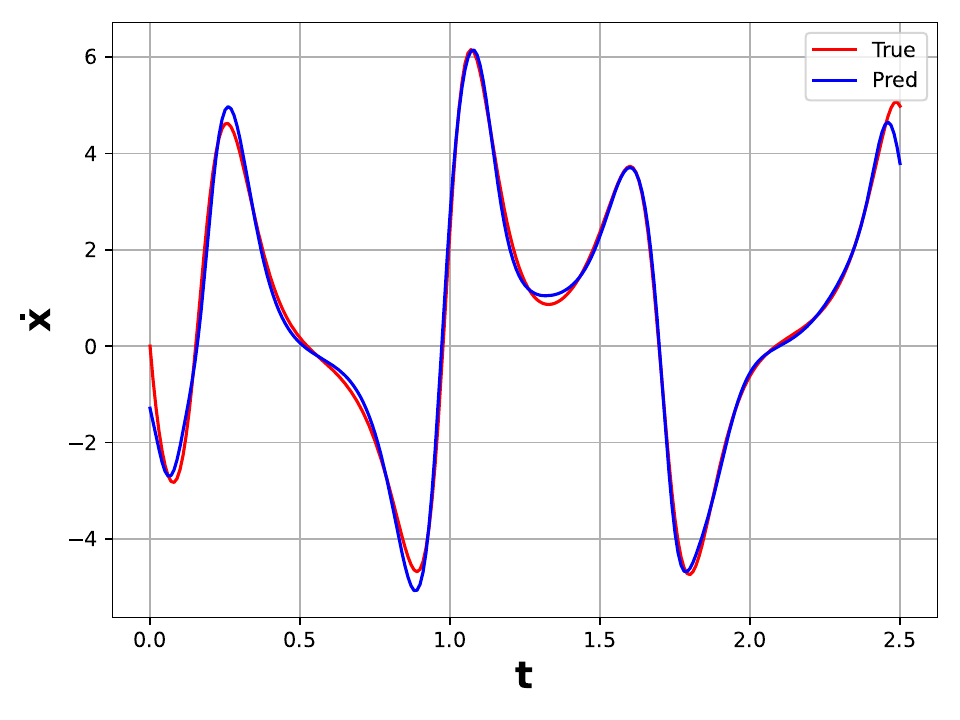}
         \caption{reference Vs predicted dynamics $\dot{x}$}
         %\label{}
     \end{subfigure}
     %\hfill
     \begin{subfigure}[b]{0.45\textwidth}
         \centering
         \includegraphics[width=.95\linewidth]{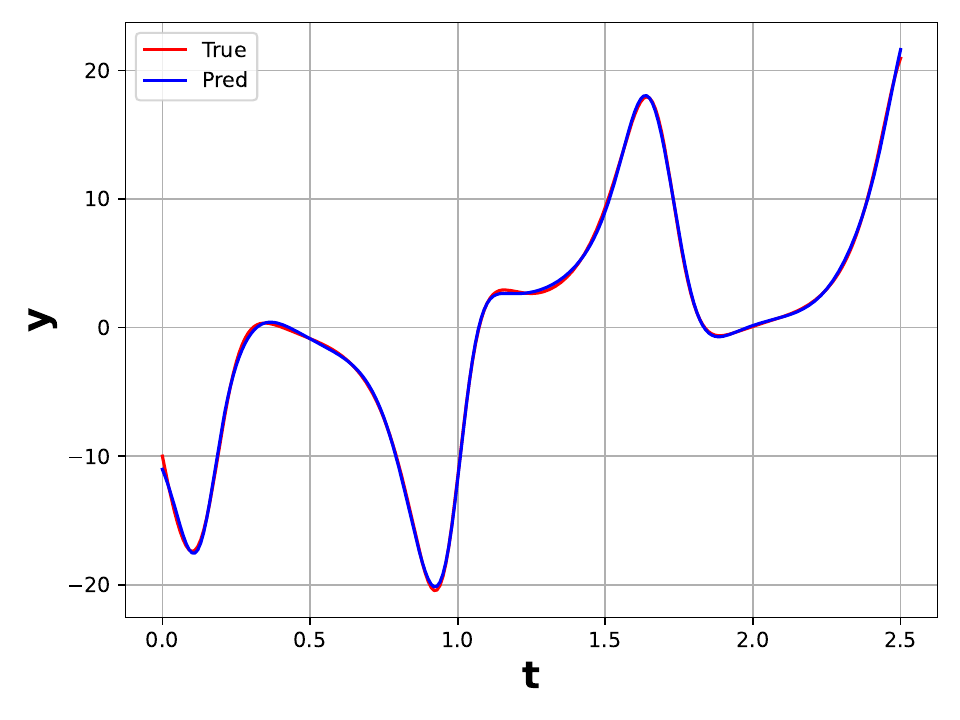}
         \caption{reference Vs predicted of state $y$ }
         %\label{}
     \end{subfigure}
     %\hfill
     \begin{subfigure}[b]{0.45\textwidth}
         \centering
         \includegraphics[width=.95\linewidth]{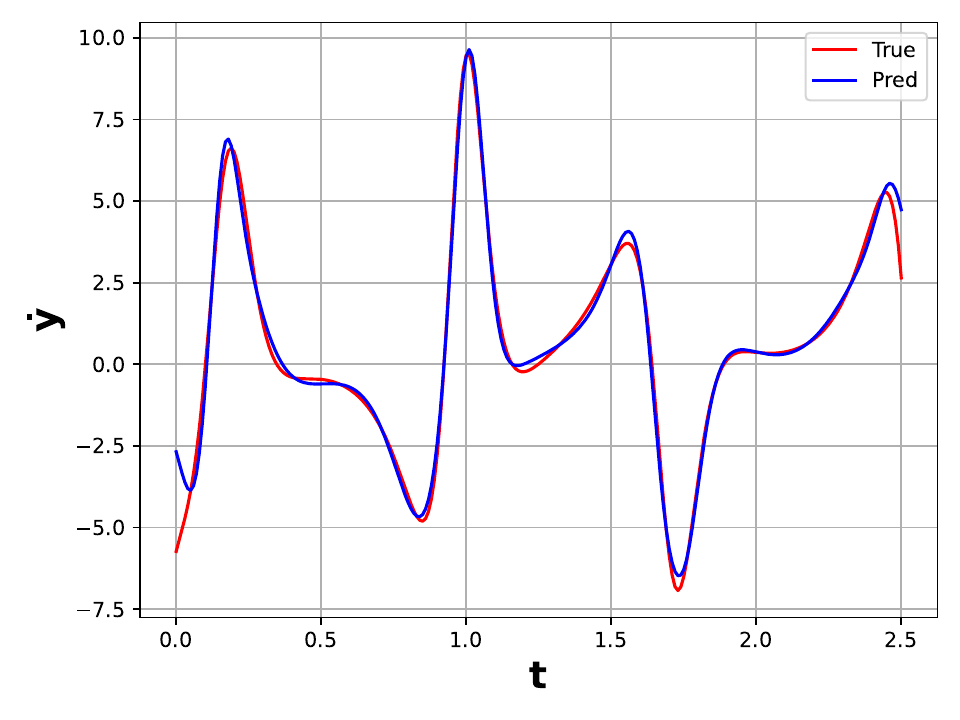}
         \caption{reference Vs predicted dynamics $\dot{y}$}
         %\label{}
     \end{subfigure}
     \begin{subfigure}[b]{0.45\textwidth}
         \centering
         \includegraphics[width=.95\linewidth]{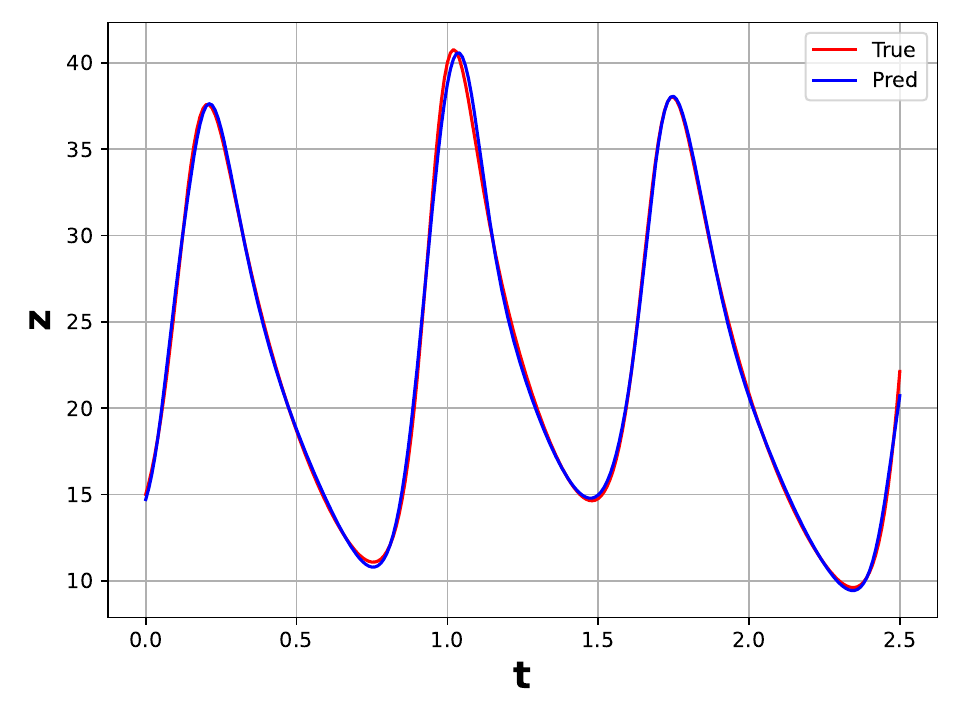}
         \caption{reference Vs predicted of state $z$ }
         %\label{}
     \end{subfigure}
     %\hfill
     \begin{subfigure}[b]{0.45\textwidth}
         \centering
         \includegraphics[width=.95\linewidth]{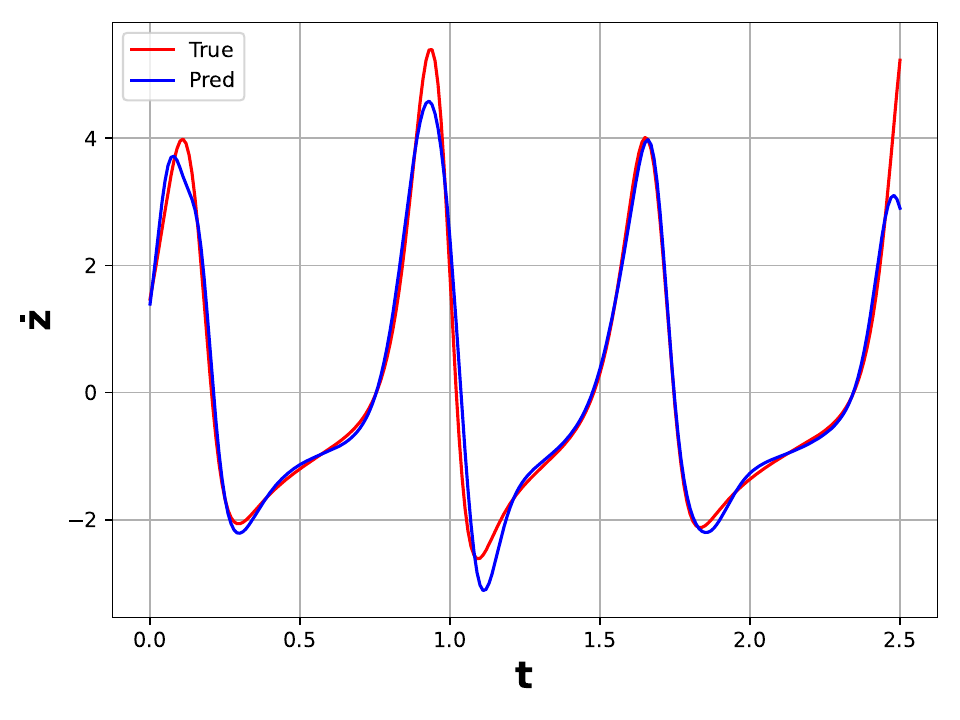}
         \caption{reference Vs predicted dynamics $\dot{z}$}
         %\label{}
     \end{subfigure}
        \caption{Lorenz system, training with $\alpha_x = 1,\alpha_y = 1,\alpha_z = 1$ }
        \label{LS_a1_1_a2_1_a3_1_sol}
\end{figure}

This shows that the data loss of all the states are contributing to net data loss. We train network with 
using 251 data points from reference solution of states $x, y, z$.
and also we consider 1000 random collocation points. The network architecture for $\mathscr{N}_s$ consists of 2 hidden layers with 128 neurons each and network $\mathscr{N}_f$ consists of 3 hidden layers with 128 neurons each. We train the model for 100000 epochs with learning rate of $10^{-3}$ and next 100000 epochs with learning rate of $10^{-4}$ until the total loss converges to order of $\mathcal{O}(10^{-4})$. The results are shown in figure  \ref{LS_a1_1_a2_1_a3_1_sol},

\subsubsection{\label{type_LS_a1_1_a2_1_a3_0} Case2: $\alpha_x = 1,\alpha_y = 1, \alpha_z = 0$ }
\begin{figure}
     \centering
     \begin{subfigure}[b]{0.45\textwidth}
         \centering
         \includegraphics[width=.95\linewidth]{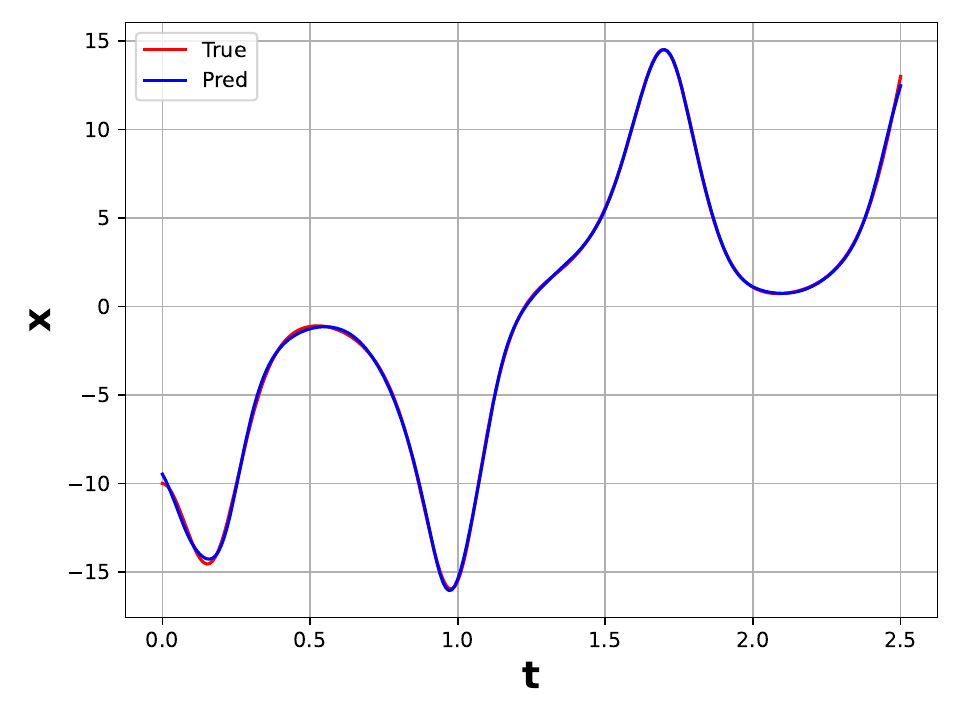}
         \caption{reference Vs predicted of state $x$ }
         %\label{}
     \end{subfigure}
     %\hfill
     \begin{subfigure}[b]{0.45\textwidth}
         \centering
         \includegraphics[width=.95\linewidth]{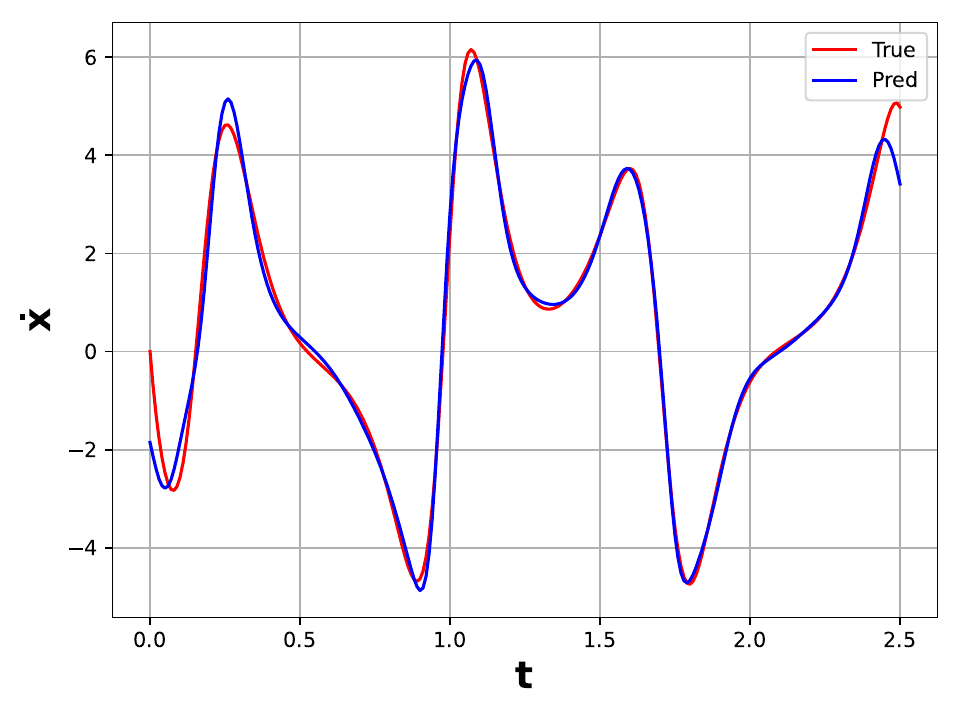}
         \caption{reference Vs predicted dynamics $\dot{x}$}
         %\label{}
     \end{subfigure}
     %\hfill
     \begin{subfigure}[b]{0.45\textwidth}
         \centering
         \includegraphics[width=.95\linewidth]{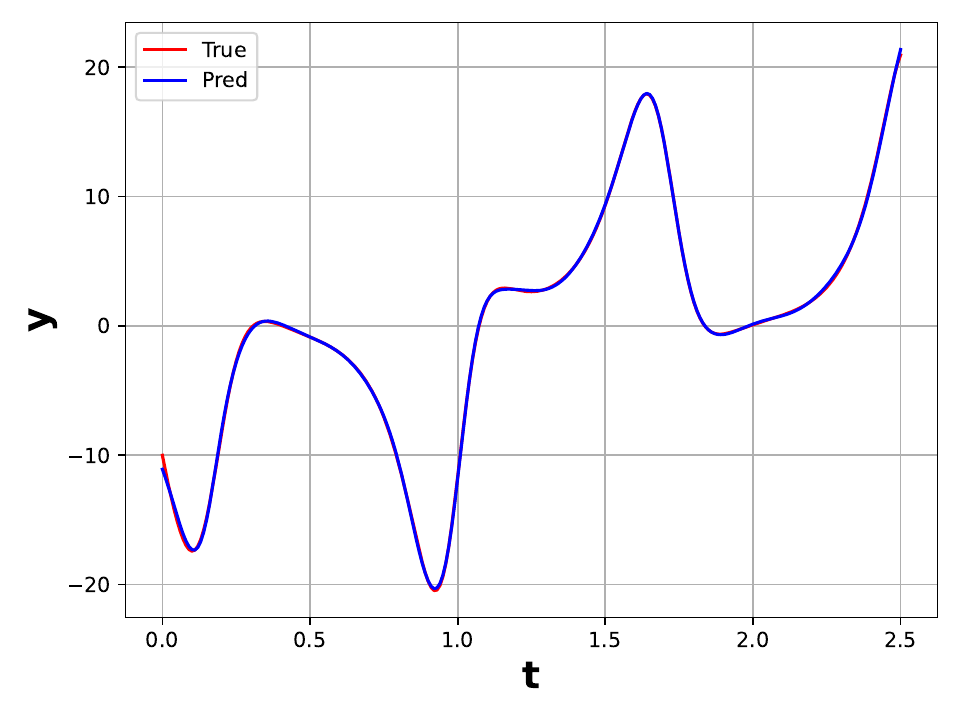}
         \caption{reference Vs predicted of state $y$ }
         %\label{}
     \end{subfigure}
     %\hfill
     \begin{subfigure}[b]{0.45\textwidth}
         \centering
         \includegraphics[width=.95\linewidth]{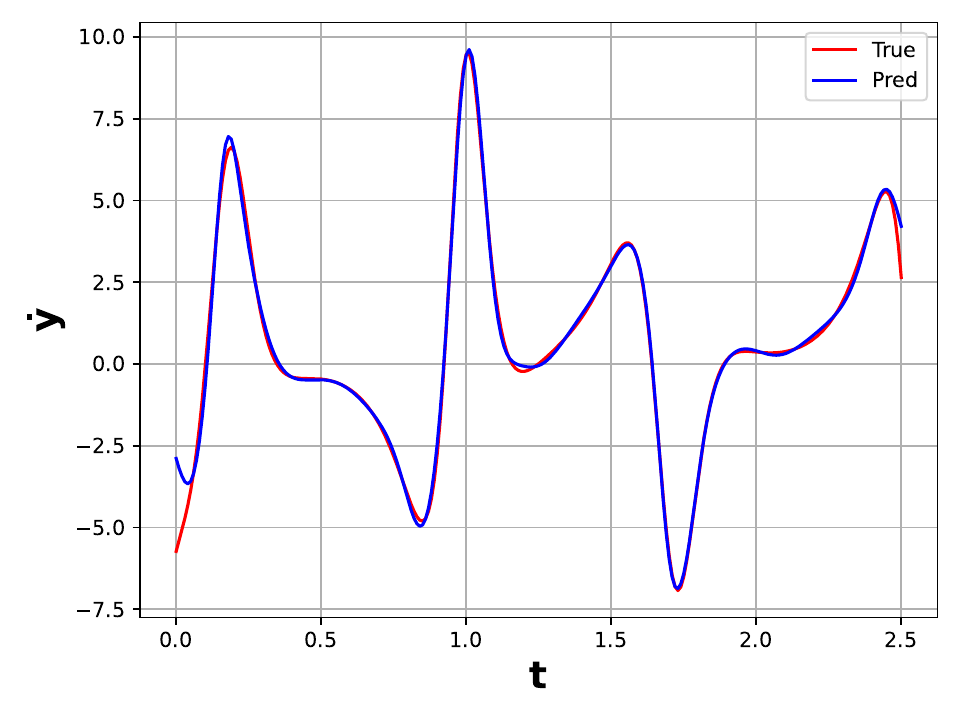}
         \caption{reference Vs predicted dynamics $\dot{y}$}
         %\label{}
     \end{subfigure}
     \begin{subfigure}[b]{0.45\textwidth}
         \centering
         \includegraphics[width=.95\linewidth]{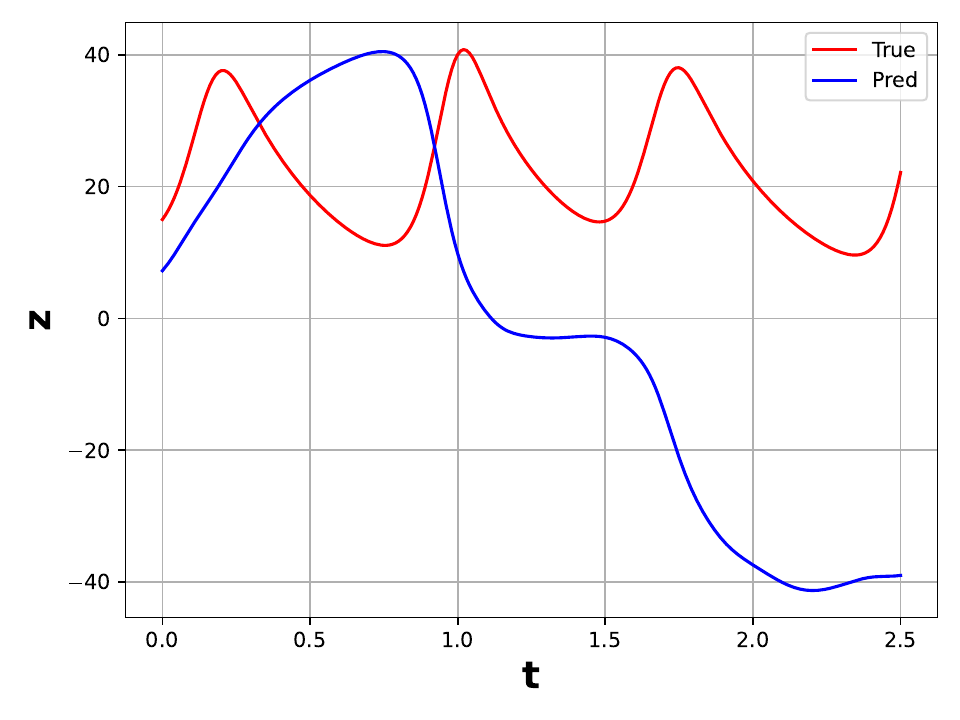}
         \caption{reference Vs predicted of state $z$ }
         %\label{}
     \end{subfigure}
     %\hfill
     \begin{subfigure}[b]{0.45\textwidth}
         \centering
         \includegraphics[width=.95\linewidth]{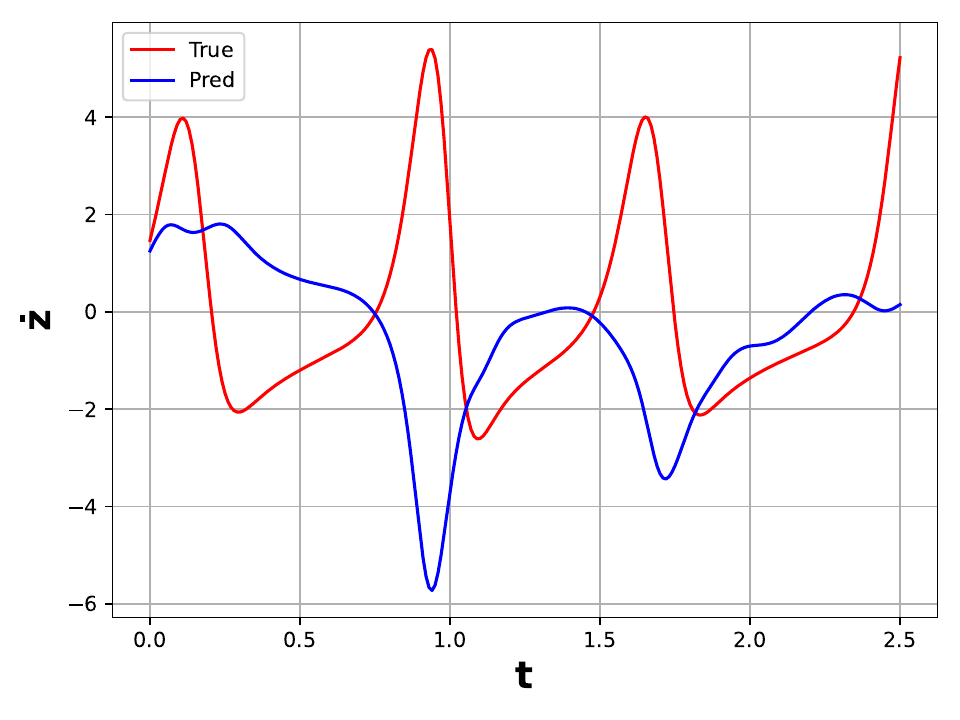}
         \caption{reference Vs predicted dynamics $\dot{z}$}
         %\label{}
     \end{subfigure}
        \caption{Lorenz system, training with $\alpha_x = 1,\alpha_y = 1,\alpha_z = 0$ }
        \label{LS_a1_1_a2_1_a3_0_sol}
\end{figure}

This shows that the data loss of only states $x$ and $y$ are contributing to net data loss. 
We train the model with same network architectures, number of data points and collocation points as shown in section \ref{type_LS_a1_1_a2_1_a3_1}
. We train the model for 100000 epochs with learning rate $10^{-3}$ and next 50000 epochs with learning rate $10^{-4}$. The results are shown in figure  \ref{LS_a1_1_a2_1_a3_0_sol}.

\subsubsection{\label{type_LS_a1_1_a2_0_a3_0} Case3: $\alpha_x = 1,\alpha_y = 0, \alpha_z = 0$ }
\begin{figure}
     \centering
     \begin{subfigure}[b]{0.45\textwidth}
         \centering
         \includegraphics[width=.95\linewidth]{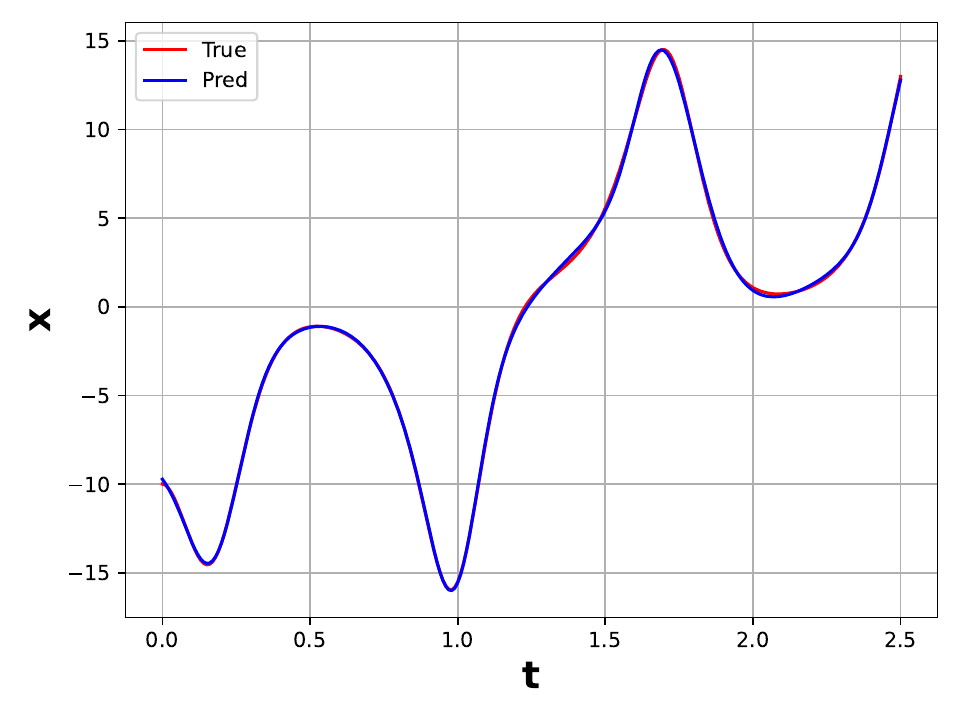}
         \caption{reference Vs predicted of state $x$ }
         %\label{}
     \end{subfigure}
     %\hfill
     \begin{subfigure}[b]{0.45\textwidth}
         \centering
         \includegraphics[width=.95\linewidth]{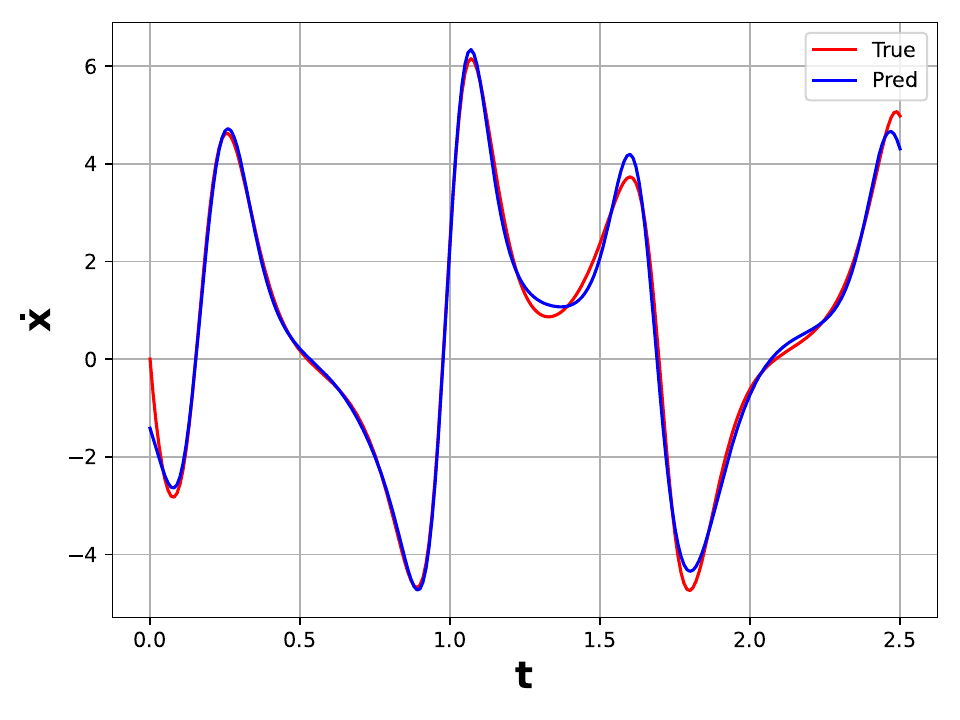}
         \caption{reference Vs predicted dynamics $\dot{x}$}
         %\label{}
     \end{subfigure}
     %\hfill
     \begin{subfigure}[b]{0.45\textwidth}
         \centering
         \includegraphics[width=.95\linewidth]{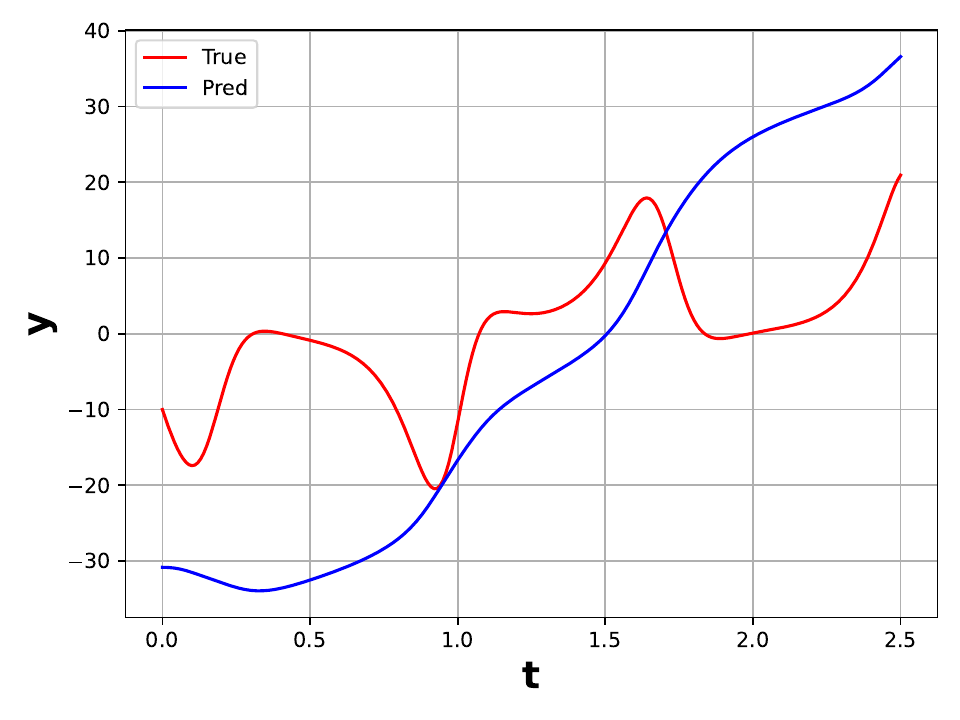}
         \caption{reference Vs predicted of state $y$ }
         %\label{}
     \end{subfigure}
     %\hfill
     \begin{subfigure}[b]{0.45\textwidth}
         \centering
         \includegraphics[width=.95\linewidth]{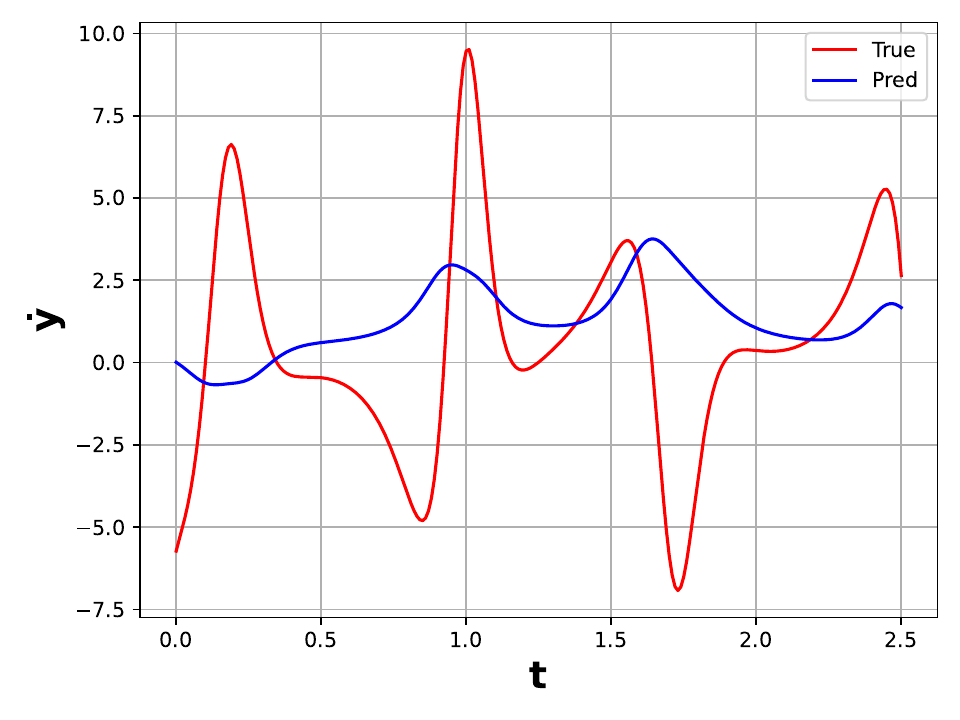}
         \caption{reference Vs predicted dynamics $\dot{y}$}
         %\label{}
     \end{subfigure}
     \begin{subfigure}[b]{0.45\textwidth}
         \centering
         \includegraphics[width=.95\linewidth]{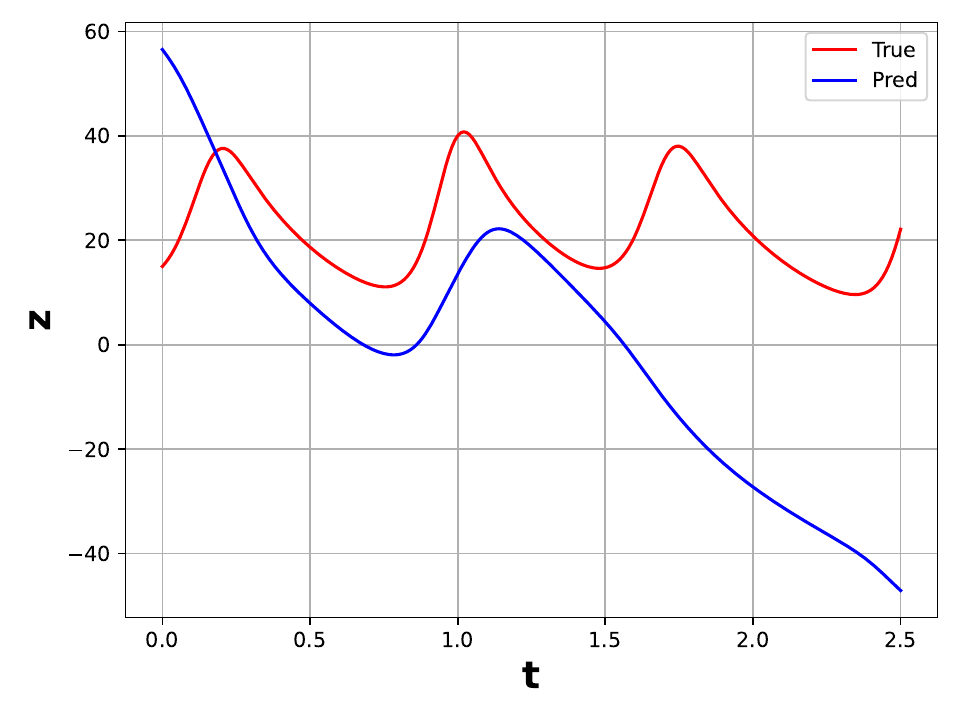}
         \caption{reference Vs predicted of state $z$ }
         %\label{}
     \end{subfigure}
     %\hfill
     \begin{subfigure}[b]{0.45\textwidth}
         \centering
         \includegraphics[width=.95\linewidth]{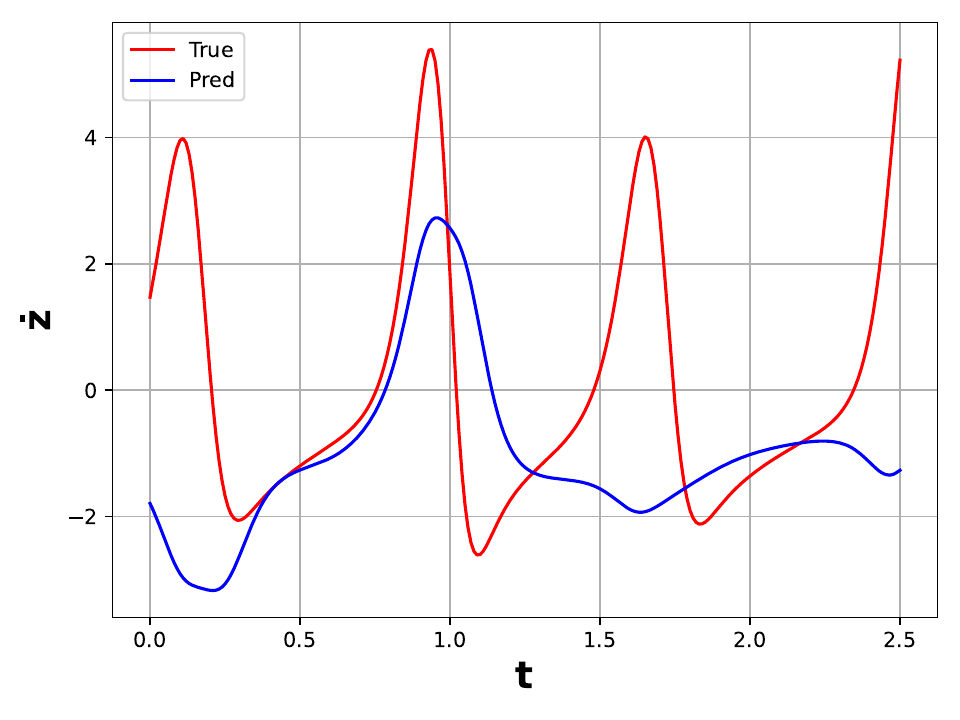}
         \caption{reference Vs predicted dynamics $\dot{z}$}
         %\label{}
     \end{subfigure}
        \caption{Lorenz system, training with $\alpha_x = 1,\alpha_y = 0,\alpha_z = 0$ }
        \label{LS_a1_1_a2_0_a3_0_sol}
\end{figure}

This shows that the data loss of only state $x$ is contributing to net data loss. We train the model with same network architectures, number of data points and collocation points as shown in section \ref{type_LS_a1_1_a2_1_a3_1}. We train the model for 100000 epochs with learning rate $10^{-3}$ and next 50000 epochs with learning rate $10^{-4}$. The results are shown in figure  \ref{LS_a1_1_a2_0_a3_0_sol}.

{\renewcommand{\arraystretch}{1.2}
\begin{table}
\caption{$L_2$ Error for Lorenz system with different contributions in data loss}
\label{LS_error}
\begin{tabular}{|l||*{3}{c|}}\hline
%\backslashbox{$\alpha$}{quantities}
&\makebox[10em]{$\alpha_x = \alpha_y = \alpha_z = 1$}&\makebox[10em]{$\alpha_x = \alpha_y = 1, \alpha_z = 0$}&\makebox[10em]{$\alpha_x = 1,\alpha_y = \alpha_z = 0$}\\\hline
$x$ & $1.62 \times 10^{-2}$ & $1.48 \times 10^{-2}$  & $1.46 \times 10^{-2}$   \\\hline
$\dot{x}$ & $8.53 \times 10^{-2}$ & $1.14 \times 10^{-1}$  & $7.67 \times 10^{-2}$  \\\hline
$y$ & $2.07 \times 10^{-2}$ & $1.39 \times 10^{-2}$  & $2.30 \times 10^{0}$   \\\hline
$\dot{y}$ & $1.25 \times 10^{-1}$ & $9.90 \times 10^{-2}$  & $1.07 \times 10^{0}$  \\\hline
$z$ & $1.62 \times 10^{-2}$ & $1.54 \times 10^{0}$  & $1.33 \times 10^{0}$   \\\hline
$\dot{z}$ & $1.61 \times 10^{-1}$ & $1.47 \times 10^{0}$  & $1.23 \times 10^{0}$  \\\hline
\end{tabular}
\end{table}
}

As We can observe in Table \ref{LS_error} that when the data of all states are given, the model is able to capture both states and dynamics perfectly with error of $\mathcal{O}(10^{-2})$. However if the data for particular state is not given in the training, the model fails to capture it's state and dynamics, however it accurately captures the dynamics of states which are given in the training.

The codes have been developed using deep learning API Tensorflow-2. The training of model is performed on NVIDIA Tesla V100 GPU. Training takes approximately 2 min for Type-a, Type-b, Nonlinear systems and 12 min for Lorenz system. As we can observe that, it takes 6X more time in training for Lorenz as compared to other systems because it takes 6X more iterations for total loss to be converged at the similar order of magnitude. 

\section{Conclusion}
We have observed in this experiments how deep hidden physics model approximates the solution and physics both based on state's data. We have considered autonomous system in our experiments therefore we only have states in candidate functions list which is then passed as input to neural network, knowing prior that the dynamics will be dependent on  states only. This fact reduces our efforts in searching for different combinations of candidate functions. We have seen in results that if we pass the data for particular states for training then the networks learns both solution and physics  accurately even though we don't consider other states data, this behaviour of neural network in learning physics without providing complete data is unanswered and we are looking for the mathematical reasons behind that.

\bibliography{main}

\begin{thebibliography}{1}

\bibitem{hirsch2004differential}
M.W. Hirsch, S.~Smale, and R.L. Devaney.
\newblock {\em Differential Equations, Dynamical Systems, and an Introduction to Chaos}.
\newblock Pure and Applied Mathematics - Academic Press. Elsevier Science, 2004.

\bibitem{adam_opt}
Diederik~P. Kingma and Jimmy Ba.
\newblock Adam: A method for stochastic optimization, 2017.

\bibitem{LHS}
M.~D. McKay, R.~J. Beckman, and W.~J. Conover.
\newblock A comparison of three methods for selecting values of input variables in the analysis of output from a computer code.
\newblock {\em Technometrics}, 21(2):239--245, 1979.

\bibitem{Raissi_dhp_2018}
Maziar Raissi.
\newblock {Deep Hidden Physics Models: Deep Learning of Nonlinear Partial Differential Equations}.
\newblock {\em Journal of Machine Learning Research}, 19(1):1--24, 2018.

\bibitem{strogatz2000nonlinear}
S.H. Strogatz.
\newblock {\em Nonlinear Dynamics and Chaos: With Applications to Physics, Biology, Chemistry and Engineering}.
\newblock Studies in nonlinearity. Westview, 2000.

\end{thebibliography}
\bibliographystyle{plain}

\newpage

\end{document}